\documentclass[10pt,journal,compsoc]{IEEEtran}

\usepackage{amsmath,amsfonts}
\usepackage{algorithmic}
\usepackage{algorithm}
\usepackage{array}
\usepackage[caption=false,font=normalsize,labelfont=sf,textfont=sf]{subfig}
\usepackage{textcomp}
\usepackage{stfloats}
\usepackage{url}
\usepackage{verbatim}
\usepackage{graphicx}
\usepackage{multirow}
\usepackage{threeparttable}
\usepackage{tikz}
\usepackage[edges]{forest}
\usepackage{bbm}
\usepackage{pbox}
\usepackage[breaklinks,colorlinks]{hyperref}
\usepackage{booktabs} 
\usepackage{fancyhdr}

\usepackage[capitalize]{cleveref}
\crefname{section}{Sec.}{Secs.}
\Crefname{section}{Section}{Sections}
\Crefname{table}{Table}{Tables}
\crefname{table}{Tab.}{Tabs.}
\crefname{equation}{Eq.}{Eqs.}

\newcommand{\ie}{\textit{i}.\textit{e}.}
\newcommand{\eg}{\textit{e}.\textit{g}.}
\newcommand{\etal}{\textit{et} \textit{al}.}
\usepackage{silence}
\usepackage{pifont}%
\newcommand{\cmark}{\ding{51}}%
\newcommand{\xmark}{\ding{55}}%
\def\eg{\emph{e.g.}}
\def\ie{\emph{i.e.}}
\def\etal{\emph{et al.}}

\def\wrt{\emph{w.r.t.~}}

\definecolor{leafcolor}{rgb}{0.8,0.8,0.8}

\ifCLASSOPTIONcompsoc
  \usepackage[nocompress]{cite}
\else
  \usepackage{cite}
\fi

\ifCLASSINFOpdf
\else
\fi

\begin{document}

\title{Learning-based Multi-View Stereo: A Survey}

\author{Fangjinhua Wang\textsuperscript{*\dag}, Qingtian Zhu\textsuperscript{*}, Di Chang\textsuperscript{*}, Quankai Gao, Junlin Han, Tong Zhang,\\
Richard Hartley,~\IEEEmembership{Fellow,~IEEE}, Marc Pollefeys\textsuperscript{\ddag},~\IEEEmembership{Fellow,~IEEE}
\IEEEcompsocitemizethanks{
\IEEEcompsocthanksitem Fangjinhua Wang and Marc Pollefeys are with the Department of
Computer Science, ETH Zurich, Switzerland. 
\IEEEcompsocthanksitem Qingtian Zhu is with the Graduate School of Information Science and Technology, The University of Tokyo, Japan.
\IEEEcompsocthanksitem Di Chang and Quankai Gao are with the Department of
Computer Science, University of Southern California, USA.
\IEEEcompsocthanksitem Junlin Han is with the Department of Engineering Science, University of Oxford, UK.
\IEEEcompsocthanksitem Tong Zhang is with the University of Chinese Academy of Sciences, China, and the School of Computer and Communication Sciences, EPFL, Switzerland.
\IEEEcompsocthanksitem Richard Hartley is with Australian National University, Australia.
\IEEEcompsocthanksitem Marc Pollefeys is additionally with Microsoft, Zurich.
}
\thanks{\textsuperscript{*}: Equal contribution. \textsuperscript{\dag}: Project lead. \textsuperscript{\ddag}: Corresponding author. }
}

\markboth{Journal of \LaTeX\ Class Files,~Vol.~14, No.~8, August~2015}%
{Shell \MakeLowercase{\textit{et al.}}: Bare Demo of IEEEtran.cls for Computer Society Journals}

\IEEEtitleabstractindextext{%
\begin{abstract}
3D reconstruction aims to recover the dense 3D structure of a scene. It plays an essential role in various applications such as Augmented/Virtual Reality (AR/VR), autonomous driving and robotics. 
Leveraging multiple views of a scene captured from different viewpoints, Multi-View Stereo (MVS) algorithms synthesize a comprehensive 3D representation, enabling precise reconstruction in complex environments. Due to its efficiency and effectiveness, MVS has become a pivotal method for image-based 3D reconstruction. 
Recently, with the success of deep learning, many learning-based MVS methods have been proposed, achieving impressive performance against traditional methods. We categorize these learning-based methods as: depth map-based, voxel-based, NeRF-based, 3D Gaussian Splatting-based, and large feed-forward methods. Among these, we focus significantly on depth map-based methods, which are the main family of MVS due to their conciseness, flexibility and scalability. In this survey, we provide a comprehensive review of the literature at the time of this writing. We investigate these learning-based methods, summarize their performances on popular benchmarks, and discuss promising future research directions in this area.
\end{abstract}

}

\maketitle

\IEEEdisplaynontitleabstractindextext

\IEEEpeerreviewmaketitle

\IEEEraisesectionheading{\section{Introduction}\label{sec:introduction}}

\IEEEPARstart{3}D reconstruction describes the general task of recovering the 3D structure of a scene. It is widely employed in augmented/virtual reality (AR/VR), autonomous driving, and robotics~\cite{furukawa2015multi}. 
The advancement of 3D acquisition techniques has led to the increased affordability and reliability of depth sensors, such as depth cameras and LiDARs. These sensors are extensively utilized for real-time tasks, enabling rough estimations of the surrounding environment, such as simultaneous localization and mapping (SLAM) ~\cite{sturm2012benchmark,endres20133,schops2019bad,pan2021mulls} or dense reconstruction ~\cite{newcombe2011kinectfusion,zollhofer2014real,newcombe2015dynamicfusion}. Nevertheless, depth maps captured by such sensors tend to be partial and sparse, resulting in incomplete 3D representations with limited geometric details. In addition, these sensors are active and usually consume a lot of power. 
In contrast, camera-based solutions, commonly found in edge devices like smartphones and AR/VR headsets, offer a more economically viable alternative for 3D reconstruction.

One fundamental technique in image-based 3D reconstruction is Multi-View Stereo (MVS). Given a set of calibrated images, MVS aims to reconstruct dense 3D geometry for an observed scene. 
Based on the scene representations, traditional MVS methods can be mainly divided into three categories: voxel, point cloud, and depth map~\cite{yao2018mvsnet}. 
Voxel-based methods~\cite{seitz1999photorealistic,kutulakos2000theory,kostrikov2014probabilistic,ulusoy2017semantic} discretize the 3D space into voxels and label each as inside or outside of the surface. They are limited to small-scale scenes due to large memory consumption. 
Point cloud-based methods~\cite{lhuillier2005quasi,furukawa2009accurate} operate directly on 3D points and often employ propagation to gradually densify the reconstruction. 
As point cloud propagation occurs sequentially, these methods are difficult to parallelize, leading to longer processing times~\cite{yao2018mvsnet}. In addition, the irregularity and large size of point cloud are also not very suitable for deep learning, especially in large-scale scenes. 
In contrast, methods that are based on depth maps~\cite{yang2003multi,schonberger2016structure,galliani2015massively,xu_2019_acmm,xu2020acmp} use patch matching with photometric consistency to estimate the depth maps for individual images. Subsequently, these 2D depth maps are fused into a dense 3D representation, \eg, point cloud or mesh. This design decouples the reconstruction task into per-view depth estimation and depth fusion, which explicitly improves flexibility and scalability. In addition, storing a set of 2D depth maps to represent 3D geometry is more convenient~\cite{seitz2006comparison,schonberger2016structure}. 
Although MVS has been studied extensively for several decades, traditional MVS methods rely on hand-crafted matching metrics and thus encounter challenges in handling various conditions, \eg, illumination changes, low-textured areas, and non-Lambertian surfaces~\cite{aanaes2016large,knapitsch2017tanks,schops2017multi,yao2020blendedmvs}.

To overcome these challenges, recent works~\cite{yao2018mvsnet,wang2018mvdepthnet} have shifted towards learning-based approaches, have adopted learning-based approaches, using convolutional neural networks (CNNs) that have excelled in 2D vision tasks. These methods have significantly outperformed traditional methods on various benchmarks~\cite{aanaes2016large,knapitsch2017tanks,schops2017multi,dai2017scannet}.

In this survey, motivated by previous works~\cite{seitz2006comparison,furukawa2009accurate,yao2018mvsnet}, we categorize learning-based MVS methods based on their underlying representations as follows: depth map-based, voxel-based, NeRF-based, 3D Gaussian Splatting-based, and large feed-forward methods. 
Voxel-based methods estimate geometry with volumetric representation and Signed Distance Functions (SDF). They are limited to small-scale scenes because of the high memory storage of voxel grids. 
NeRF and 3D Gaussian Splatting-based methods adapt NeRF~\cite{mildenhall2020nerf} and 3D Gaussian Splatting~\cite{kerbl3Dgaussians}, which are used for novel view synthesis, and extract the surface from the volumetric implicit field and explicit 3D Gaussians points, respectively. They typically need to optimize the geometry for each scene. 
Large feed-forward methods typically use large models to directly learn the 3D representation from given images. They require massive computation because of the large network. 
Depth map-based methods introduce deep learning in depth estimation and then fuse depth maps with the traditional fusion algorithm. 
Similar to traditional MVS, depth map-based methods~\cite{yao2018mvsnet,wang2018mvdepthnet} are the main family of learning-based MVS since they inherit the advantages of those traditional methods~\cite{schonberger2016pixelwise} by decoupling 3D reconstruction into depth estimation and depth fusion. 
Therefore, we focus more on depth map-based MVS in this survey. 

\begin{figure*}
    \centering
    \includegraphics[width=\linewidth]{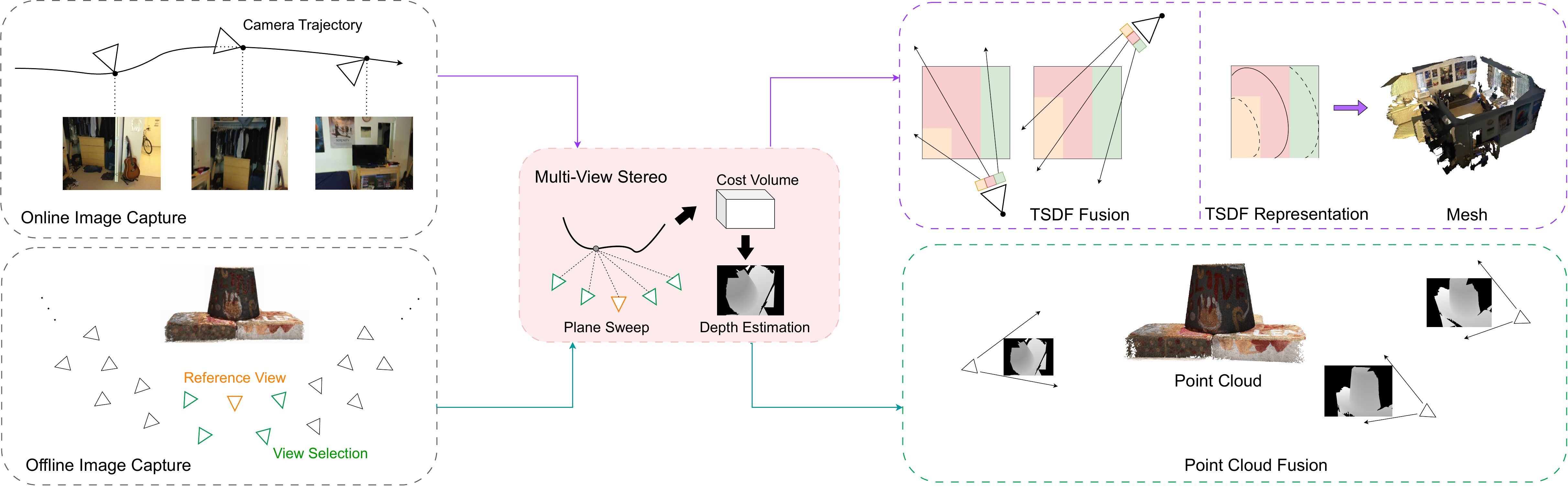}
    \caption{An overall illustration of both online and offline depth map-based MVS pipelines. Online MVS usually deals with sequential data, \eg, video, and employs TSDF volumes as an intermediate representation for mesh extraction. Given a full set of images, offline MVS holds the global information of the captured scene, and usually fuses estimated depth maps into a point cloud with filtering. 
    }
    \label{fig:overall}
\end{figure*}

For clarity, we further categorize depth map-based MVS into \textit{online} and \textit{offline} methods, shown in \cref{fig:overall}. 
Online methods~\cite{wang2018mvdepthnet,duzceker2021deepvideomvs} reconstruct scenes in an online manner with local temporal information, while offline methods~\cite{yao2018mvsnet,gu2020cascade} reconstruct scenes in an offline manner with global information. To estimate depth for a given reference view, online methods select its best views in a local buffer that stores previous views. In contrast, offline methods select best views among all possible views in the database.

In summary, our survey covers the most recent literature on learning-based MVS methods,  %
including five main families: depth map-based, voxel-based, NeRF-based, 3D Gaussian Splatting-based, and large feed-forward methods. %
We provide a comprehensive review and insights on different aspects, including the pipelines and algorithmic intricacies. 
Moreover, we evaluate the performance and efficiency of the reviewed methods, and discuss the potential future directions for deep learning-based MVS. 

\section{Preliminaries}\label{sec:background}
Depth map-based MVS, including most traditional and learning-based methods, typically consists of several components: camera calibration, view selection, multi-view depth estimation and depth fusion. 
In this section, we introduce these components to provide readers a clear picture of the MVS. Note that camera calibration and view selection are generally required by other methods as well. %
\cref{sec:sfm} introduces camera calibration with Structure from Motion (SfM) or SLAM. 
\cref{sec:viewselect} discusses how to select neighboring views to  reconstruct the geometry. 
\cref{sec:ps} explains how to build cost volumes in learning-based MVS methods with plane sweep~\cite{collins1996space}. %
\cref{sec:dff} introduces the typical depth fusion strategies after depth estimation. 
\cref{sec:datasets} lists common datasets and benchmarks for MVS and \cref{sec:em} summarizes common evaluation metrics.

\tikzstyle{leaf}=[draw=black,
    rounded corners,minimum height=1.2em,
    fill=leafcolor,
    text opacity=1, align=center,
    fill opacity=.5,  text=black,align=left,font=\scriptsize,
inner xsep=3pt,
inner ysep=1pt,
]
\begin{figure*}[t]
\centering
\begin{forest}
  for tree={
  forked edges,
  grow=east,
  reversed=true,
  anchor=base west,
  parent anchor=east,
  child anchor=west,
  base=middle,
  font=\footnotesize,
  rectangle,
  draw=black,
  rounded corners,align=left,
  minimum width=2.5em,
  minimum height=1.2em,
    s sep=6pt,
    inner xsep=3pt,
    inner ysep=1pt,
  },
  where level=1{text width=4.5em}{},
  where level=2{text width=8em,font=\scriptsize}{},
  where level=3{font=\scriptsize}{},
  where level=4{font=\scriptsize}{},
  where level=5{font=\scriptsize}{},
  [\textbf{MVS}
    [Traditional Methods,text width=10 em
        [COLMAP~\cite{schonberger2016pixelwise}{,}
        P-MVS~\cite{furukawa2009accurate}{,}
        Gipuma~\cite{galliani2015massively}...,leaf,text width=16.3 em]
    ]
    [Datasets \& Benchmarks, text width=10 em
        [ScanNet~\cite{dai2017scannet}{,}
        DTU~\cite{aanaes2016large}{,}
        Tanks and Temples~\cite{knapitsch2017tanks}...,
        leaf,text width=16.3 em]
    ]
    [Learning-based MVS \\ with Depth Representation,text width=11 em
        [Pipeline,text width=6 em
            [Camera Calibration{,} View Selection{,} Multi-View Depth Estimation{,} Depth Fusion,leaf,text width=27.4 em]
        ]
        [Supervised,text width=6em
            [Online Methods
                [MVDepthNet~\cite{wang2018mvdepthnet}{,} DeepvideoMVS~\cite{duzceker2021deepvideomvs}{,} SimpleRecon~\cite{sayed2022simplerecon}...,leaf,text width=20.2 em]
            ]
            [Offline Methods
                [Direct 3D CNN,text width=5.4 em
                    [MVSNet~\cite{yao2018mvsnet}{,}
                   CIDER~\cite{xu2020learning_inverse}...,leaf,text width=13 em]
                ]
                [RNN,text width=5.4 em
                    [R-MVSNet~\cite{yao2019recurrent}{,}
                    $D^2$HC-RMVSNet~\cite{yan2020dense}...,leaf,text width=13 em]
                ]
                [Coarse-to-fine,text width=5.4 em
                    [CasMVSNet~\cite{gu2020cascade}{,}
                    UCSNet~\cite{cheng2020deep}...,leaf,text width=13 em]
                ]
                [Iterative Update,text width=5.4 em
                    [PatchmatchNet~\cite{wang2021patchmatchnet}{,}
                    IterMVS~\cite{wang2021itermvs}...,leaf,text width=13 em]
                ]
            ]
        ]
        [Unsupervised,text width=6em
            [End-to-end,text width=4 em
                [
                JDACS~\cite{xu2021self}{,}
                RC-MVSNet~\cite{chang2022rc}...,leaf,text width=10.5 em]
            ]
            [Multi-stage,text width=4 em
                [%
                U-MVSNet~\cite{xu2021digging}{,}
                KD-MVS~\cite{ding2022kdmvs}...,leaf,text width=10.5 em]
            ]
        ]
        [Semi-supervised,text width=6em
          [
            SGT-MVSNet~\cite{kim2021just},leaf,text width=6.2 em]
          ]
        ]
    [Learning-based MVS \\ with Other Representations,text width=11em
        [Voxel,text width=8 em
            [Direct 3D CNN,text width=6 em
                [
                Atlas~\cite{murez2020atlas},leaf,text width=15 em]
            ]
            [Coarse-to-fine,text width=6 em
                [NeuralRecon~\cite{sun2021neuralrecon}{,}
                TransformerFusion~\cite{bozic2021transformerfusion}...,leaf,text width=15 em]
            ]     
        ]
        [NeRF,text width=8 em
            [Optimization,text width=5 em
                [NeuS~\cite{wang2021neus}{,}
                Neuralangelo~\cite{li2023neuralangelo}{,}
                UniSDF~\cite{wang2023unisdf}...,leaf,text width=16 em]
            ]
            [Generalizable,text width=5 em
                [SparseNeuS~\cite{long2022sparseneus}{,}
                VolRecon~\cite{ren2022volrecon}{,}
                LRM~\cite{hong2023lrm}...,leaf,text width=16 em]
            ]     
        ]
        [3D Gaussian Splatting,text width=8 em
            [Optimization,text width=5 em
                [
                SuGaR~\cite{guedon2023sugar}{,}
                2DGS~\cite{huang20242d}{,}
                PGSR~\cite{chen2024pgsr}...,leaf,text width=16 em]
            ]
            [Generalizable,text width=5 em
                [MVSplat~\cite{chen2024mvsplat}{,}
                DepthSplat~\cite{xu2025depthsplat}{,}
                LGM~\cite{tang2024lgm}...,leaf,text width=16 em]
            ]     
        ]
        [Large Feed-forward Point-based,text width=15 em
            [DUSt3R~\cite{wang2023dust3r}{,}
                VGGT~\cite{wang2025vggt}{,}
                MapAnything~\cite{keetha2025mapanything}...,leaf,text width=16 em] 
        ]
    ]
  [Future Directions, text width=7 em
  [Dataset \& Benchmarks{,}
  View Selection{,}
  Feature Extraction{,}
  Generative Reconstruction{,}
  Efficiency{,}
  Prior Assistance,
  leaf,text width=38 em]
  ]
    ]
\end{forest}
\caption{Taxonomy of Multi-View Stereo.}
\label{fig:taxonomy}
\end{figure*}

\subsection{
Camera Calibration
}\label{sec:sfm}

Camera calibration is a process of determining the intrinsic and extrinsic parameters of a camera to understand its geometry and characteristics accurately~\cite{hartley2003multiple}. It serves as the foundational step in MVS, ensuring that the subsequent reconstruction process is built on accurate and consistent geometric information, ultimately leading to a more reliable and precise 3D representation of the scene. %
Typically, obtaining calibrated camera parameters is usually achieved by running off-the-shelf SfM algorithms~\cite{schonberger2016structure, moulon2016openmvg} or SLAM~\cite{dai2017bundlefusion}, which jointly optimize sparse triangulated 3D points and camera parameters. The camera parameters include the extrinsic matrix $\mathbf{T} = [\mathbf{R} | \mathbf{t}]$ and intrinsic matrix $\mathbf{K}$. 
Depth map-based MVS methods~\cite{yao2018mvsnet,gu2020cascade,wang2021patchmatchnet} require a bounded depth range $[d_{\textrm{min}},d_{\textrm{max}}]$ to improve the estimation accuracy. For offline methods~\cite{yao2018mvsnet,gu2020cascade}, the depth range can be estimated by projecting the sparse point cloud from SfM to each viewpoint and compute the minimum and maximum $z$ values~\cite{yao2018mvsnet}. 
In contrast, online methods~\cite{wang2018mvdepthnet,duzceker2021deepvideomvs} usually set constant depth ranges, \eg, [0.25$m$, 20.00$m$], since the scene scale is usually fixed as room. 

\subsection{View Selection}\label{sec:viewselect}
The selection of views is an important step for reconstruction. 
It is important to balance triangulation quality, matching accuracy, and view frustum overlap\cite{duzceker2021deepvideomvs}. 
Currently, there are two main strategies for view selection. 

First, for most online MVS  methods~\cite{wang2018mvdepthnet,hou2019multi,duzceker2021deepvideomvs,sayed2022simplerecon}, a frame is selected as a keyframe when its pose has sufficient difference compared with the previous keyframe. Within a keyframe buffer, each new keyframe selects several old keyframes with smallest pose difference to estimate depth~\cite{duzceker2021deepvideomvs}.  %
To measure this pose difference, GP-MVS~\cite{hou2019multi} proposes a heuristic measure as:
\begin{equation}
    \text{disc}(\mathbf{T}_{ij}) = \sqrt{||\mathbf{t}_{ij}||^2 + \frac{2}{3} \text{tr} (\mathbbm{I}-\mathbf{R}_{ij})},
\end{equation}
where $\mathbf{T}_{ij} = [\mathbf{R}_{ij} | \mathbf{t}_{ij}]$ is the relative transformation between view $i$ and $j$. A new keyframe is added if the pose-distance $\text{disc}(\mathbf{T}_{ij})$ from the most recent keyframe is above a certain threshold~\cite{duzceker2021deepvideomvs,sayed2022simplerecon}. 

Second, for most offline MVS methods~\cite{yao2018mvsnet,yao2019recurrent,gu2020cascade}, view selection is done with the sparse point cloud obtained by SfM~\cite{schonberger2016pixelwise, moulon2016openmvg}. 
For a reference view $i$, MVSNet~\cite{yao2018mvsnet} computes a score $s(i,j) = \sum_{\mathbf{P}} \eta(\mathbf{\Theta}_{ij}(\mathbf{P}))$ for the neighboring view $j$, where $\mathbf{P}$ is a 3D point observed by both view $i$ and $j$. 

$\mathbf{\Theta}_{ij}(\mathbf{P})$ is the baseline angle for $\mathbf{P}$, which is computed as:
\begin{equation}
    \mathbf{\Theta}_{ij}(\mathbf{P}) = (180 / \pi) \arccos{((\mathbf{c}_i)-\mathbf{P}) \cdot ((\mathbf{c}_j)-\mathbf{P}))},
\end{equation}
where $\mathbf{c}_i, \mathbf{c}_j$ are the camera centers. 
$\eta(\cdot)$ is a piece-wise Gaussian function~\cite{zhang2015joint} to favor a certain baseline angle $\theta_0$:
\begin{equation}
    \eta(\theta) = \left\{
    \begin{aligned}
        \exp{(-\frac{(\theta - \theta_0)^2}{2\sigma_1^2})}, \theta \leq \theta_0 \\
        \exp{(-\frac{(\theta - \theta_0)^2}{2\sigma_2^2})}, \theta > \theta_0
    \end{aligned},
    \right.
\end{equation}
where $\theta_0$ is the favored baseline angle, and $\sigma_0$ and $\sigma_1$ are weights. 
Then the view selection is done by choosing neighboring views with highest scores. Almost all the following offline MVS methods~\cite{yao2019recurrent,gu2020cascade,wang2021patchmatchnet} use the same strategy.

\subsection{Multi-view Depth Estimation with Plane Sweep}\label{sec:ps}

To form a structured data format that is more suitable for convolution operations, most depth map-based MVS methods rely on the plane sweep algorithm~\cite{collins1996space} to calculate matching costs. The plane sweep algorithm discretizes the depth space with a set of fronto-parallel planes along the depth direction. 
This practice is deeply inspired by learning-based binocular stereo methods~\cite{kendall2017end,psmnet,guo2019group}, which assess matching costs for a set of disparity hypotheses and subsequently estimate the disparity.

In a nutshell, the plane sweep algorithm entails iteratively sweeping planes through the object space, computing homographies between images, and selecting depth values based on consensus among different views, ultimately facilitating accurate 3D reconstruction. 
The plane sweep algorithm discretizes the depth space using a series of parallel planes along the depth direction. It operates by sweeping a conceptual plane through the object space and evaluating the spatial distribution of geometric surfaces.

In practice, we divide the depth range, which is manually set~\cite{wang2018mvdepthnet} or estimated by SfM~\cite{yao2018mvsnet}, into discrete samples and assign hypotheses at these values. 
To map coordinates with depth hypothesis $d$, homography transformation~\cite{yao2018mvsnet} $\mathbf{p}_{i}(d) \sim \mathbf{H}_i(d) \cdot \mathbf{p}$, is applied, where $\mathbf{p}_{i}(d)$ is the corresponding pixel in the $i$-th source view for the pixel $\mathbf{p}$ of the reference view.
The homography $\mathbf{H}_i(d)$ between the $i$-th source view and the reference view 0 can be computed as: 
\begin{equation}\label{equ:homo}
    \mathbf{H}_i(d) = \mathbf{K}_i \mathbf{R}_i \left(\mathbbm{I} - \frac{(-\mathbf{R}_0^{\top}\mathbf{t}_0+\mathbf{R}_i^{\top}\mathbf{t}_i)  \mathbf{n}^{\top} \mathbf{R}_0}{d}\right) \mathbf{R}_0^{\top} \mathbf{K}_0^{-1},
\end{equation}
where $\mathbf{K}_0$, $\mathbf{K}_i$ denote camera intrinsics, $[\mathbf{R}_{0}|\mathbf{t}_{0}]$, $[\mathbf{R}_{i}|\mathbf{t}_{i}]$ denote camera extrinsics, $\mathbf{n}$ denotes the principal axis of the reference view. 
Equivalently, we can also project a reference pixel in the source views with depth hypothesis~\cite{xu2020learning_inverse,wang2021patchmatchnet,wang2021itermvs}. %
We compute $\mathbf{p}_{i}(d)$ in the $i$-th source view for pixel $\mathbf{p}$ in the reference and depth $d$ as follows:

\begin{equation}\label{eq:point_project}
\setlength{\abovedisplayskip}{3pt}
\setlength{\belowdisplayskip}{6pt}
    \mathbf{p}_{i}(d) = \mathbf{K}_i \cdot \left(\mathbf{R}_{i} \mathbf{R}_{0}^{\top} \cdot(\mathbf{K}_0^{-1} \cdot \mathbf{p} \cdot d) - \mathbf{R}_{i} \mathbf{R}_{0}^{\top} \cdot \mathbf{t}_0 + \mathbf{t}_i \right).
\end{equation}
The warped source feature is then obtained via differentiable bilinear interpolation of the source feature map. 
To better elaborate the process of plane sweep, we include a toy example in the supplement. 
The photometric similarity (or matching cost) is measured in a one-to-many manner between reference and warped source features for the following depth estimation, which will be discussed in \cref{sec:cvc}.

\subsection{Depth Fusion}\label{sec:dff}
For depth map-based MVS, after estimating all the depth maps, we need to fuse them into a dense 3D representation, \eg, point cloud or mesh. 
Online MVS methods~\cite{duzceker2021deepvideomvs,sayed2022simplerecon} usually adopt TSDF (Truncated Signed Distance Function) fusion~\cite{curless1996volumetric,newcombe2011kinectfusion} to sequentially fuse depth maps into a TSDF volume and then use Marching Cube~\cite{lorensen1998marching} to extract the mesh. 
However, there usually exist outliers in the depth maps, which may reduce the accuracy of reconstruction. 
To adress this problem, offline MVS methods~\cite{yao2018mvsnet,gu2020cascade,wang2021patchmatchnet} filter out
outliers before converting the depth maps to point clouds. 
Depth filtering involves photometric and geometric filtering~\cite{yao2018mvsnet} (for more details, please refer to the supplement). Note that online MVS methods do not perform depth filtering before TSDF fusion. For depth fusion, offline MVS methods use the visibility-based fusion algorithm~\cite{merrell2007real} to minimize depth occlusions and violations across different viewpoints. The fused depth maps are then directly reprojected to space to generate the 3D point cloud~\cite{yao2018mvsnet}.

\subsection{Datasets and Benchmarks}\label{sec:datasets}

High-quality and large-scale 3D datasets and benchmarks are important for both training and evaluation. We briefly summarize the commonly used MVS datasets and benchmarks in ~\cref{tab:datasets}.

\noindent \textbf{SUN3D}~\cite{xiao2013sun3d} is a RGB-D video dataset for large indoor scenes, which provides RGB-D images, camera poses, segmentations, and point clouds.

\noindent \textbf{ScanNet}~\cite{dai2017scannet} is a large RGB-D dataset that contains 1613 indoor scenes with ground-truth camera poses, depth maps, surface reconstruction, and semantic segmentation labels. 
Online MVS methods~\cite{wang2018mvdepthnet,duzceker2021deepvideomvs,sayed2022simplerecon} mainly use ScanNet for training and testing. 

\noindent \textbf{7-Scenes}~\cite{glocker2013real} is a RGB-D dataset captured in indoor scenes with a handheld Kinect RGB-D camera. Since it is relatively small, 7-Scenes is usually used to test the generalization performance of the models trained on ScanNet~\cite{dai2017scannet} without finetuning~\cite{duzceker2021deepvideomvs}.

\noindent \textbf{DTU}~\cite{aanaes2016large} is an object-centric MVS dataset collected under well-controlled laboratory conditions with known camera trajectory. It contains 128 scans with 49 or 64 views under 7 different lighting conditions. %
Since DTU dataset provides scanned ground truth point clouds instead of depth maps, it is required to generate mesh models with surface reconstruction, \eg, screened Poisson surface reconstruction algorithm~\cite{kazhdan2013screened}, and then render depth maps~\cite{yao2018mvsnet} for training.

\noindent \textbf{Tanks and Temples}~\cite{knapitsch2017tanks} is a large-scale benchmark captured in more complex real indoor and outdoor scenarios. It is divided into intermediate and advanced sets.  Different scenes have different scales, surface reflection and exposure conditions. %
Note that Tanks and Temples does not provide ground truth camera parameters, which are usually estimated with Structure-from-Motion~\cite{schonberger2016structure,moulon2016openmvg}. %

\noindent \textbf{ETH3D}~\cite{schops2017multi} contains 25 large-scale indoor and outdoor scenes with high-resolution RGB images. 
The scenes typically contain many low-textured regions and non-Lambertian surfaces, \eg, white walls and reflective floor. In addition, the images are usually sparse and have strong viewpoint variations and occlusions. Therefore, ETH3D is considered as a very challenging benchmark. %

\noindent \textbf{BlendedMVS}~\cite{yao2020blendedmvs} is a large-scale synthetic dataset for MVS training. It contains a variety of scenes, such as cities, sculptures, and shoes. The dataset consists of more than 17$k$ high-resolution images rendered with reconstructed models and is split into 106 training scenes and 7 validation scenes. Since images are rendered through virtual cameras, the camera parameters are accurate enough for training. 

\noindent \textbf{ScanNet++}~\cite{yeshwanth2023scannet++} is a large-scale and high-resolution indoor dataset with 3D reconstructions, high-quality
RGB images, commodity RGB-D video. Compared to ScanNet~\cite{dai2017scannet}, the quality of color and geometry is higher.

\begin{table*}[t!]
    \caption{An overview of commonly used datasets and benchmarks for learning-based Multi-View Stereo. %
    }
    \centering
    \resizebox{\textwidth}{!}{ 
    \begin{threeparttable}
    {
    \begin{tabular}{l|c|c|c|c|c|c|c|c|c}
    \hline
    \multirow{2}{*}{\textbf{Dataset}} & \multirow{2}{*}{\textbf{Year}} & \textbf{Num.} & \multicolumn{2}{c|}{\textbf{Scene Complexity}} & \multicolumn{4}{c|}{\textbf{Provided Ground Truth}}  & \textbf{Evaluation} \\ 
    \cline{4-9}
    & & \textbf{Scenes} & Indoor & Outdoor & Camera Pose & Depth & Point Cloud & Mesh &  \textbf{Target}\\
    \hline
    SUN3D~\cite{xiao2013sun3d}  & 2013 & 415 & \cmark & & \cmark & \cmark & \cmark & & Depth \\ 
    \hline
    7-Scenes~\cite{glocker2013real} & 2013 & 7 & \cmark & & \cmark & \cmark & & &  Depth\\ 
    \hline
    DTU~\cite{aanaes2016large} & 2016 & 124 & \cmark & & \cmark &  & \cmark & & Point Cloud \\
    \hline
    ScanNet~\cite{dai2017scannet} & 2017 & 1503 & \cmark & & \cmark & \cmark & & \cmark & Depth / Mesh \\ 
    \hline
    Tanks and Temples~\cite{knapitsch2017tanks} & 2017 & 21 & \cmark & \cmark &  &  & \cmark & &  Point Cloud\tnote{1} \\
    \hline
    ETH3D~\cite{schops2017multi} & 2017 & 25 & \cmark & \cmark & \cmark  &   & \cmark & & Point Cloud\tnote{1} \\
    \hline
    BlendedMVS~\cite{yao2020blendedmvs} & 2020 & 113 & \cmark & \cmark & \cmark & \cmark & & & Depth\\ 
    \hline
    ScanNet++~\cite{yeshwanth2023scannet++} & 2023 & 460 & \cmark & & \cmark & \cmark & & \cmark & Depth / Mesh\\ 
    \hline
    \end{tabular}
    \begin{tablenotes}
     \item[1] For datasets with online benchmark, the point cloud ground truth of test set is not released.
   \end{tablenotes}
    }
    \end{threeparttable}
    }
\label{tab:datasets}
\end{table*}

\subsection{Evaluation Metrics}\label{sec:em}
Based on the ground truth, \eg, depth maps or point clouds, evaluation can be categorized into 2D and 3D metrics. 

\noindent \textbf{2D Metrics:}
2D metrics are used to evaluate depth maps~\cite{eigen2014depth,sinha2020deltas}. The common metrics are: mean absolute depth error (Abs), mean absolute relative depth error (Abs Rel), %
and inlier ratio with threshold 1.25 ($\delta<1.25$). For more details, please refer to the supplement.

\noindent \textbf{3D Metrics:}
3D metrics are used to evaluate the accuracy of point cloud.
Note that %
the reconstructed point clouds should be aligned with ground truth point clouds before evaluation, \eg, by Iterative Closest Point (ICP), if their extrinsics are differently calibrated. The metrics are Precision/Accuracy, Recall/Completeness and F-score. For more details, please refer to the supplement.

\section{Supervised MVS with Depth Representation}\label{sec:methods}
\begin{figure*}[t!]
    \centering
    \includegraphics[width=\linewidth]{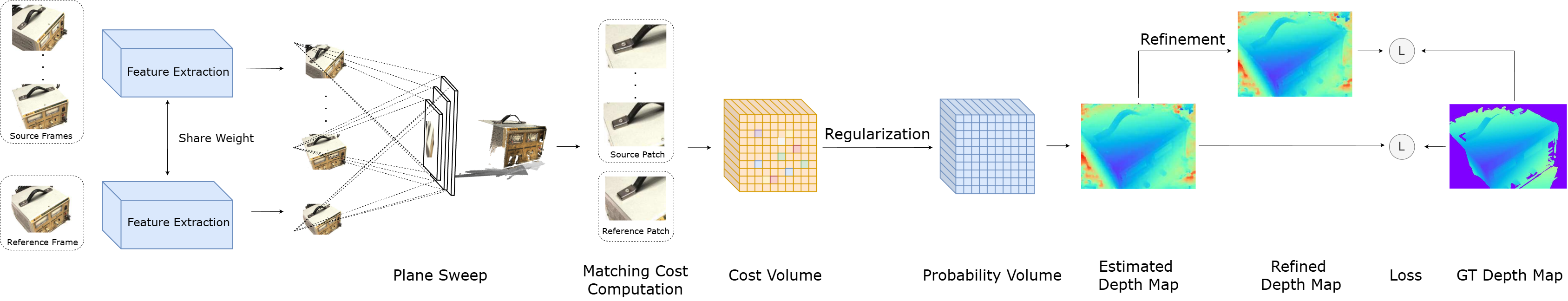}
    \vspace{-5mm}
    \caption{Pipeline of depth map-based MVS, which usually consists of feature extraction, cost volume construction via plane sweep, cost volume regularization, depth estimation, and depth refinement. }
    \label{fig:depth_mvs}
\end{figure*}

This section mainly introduces supervised MVS methods with depth representation. %
As shown in \cref{fig:depth_mvs}, a typical depth map-based MVS pipeline mainly consists of feature extraction (\cref{sec:fe}), cost volume construction (\cref{sec:cvc}), cost volume regularization (\cref{sec:cvr}), and depth estimation (\cref{sec:depth_est}).

\subsection{Feature Extraction}\label{sec:fe}
Considering efficiency, most methods use simple CNN structures to extract deep features $\left\{\mathbf{F}_i\right\}_{i=0}^{N-1}$ from images, \eg, ResNet~\cite{he2016deep}, U-Net~\cite{ronneberger2015u} and FPN~\cite{lin2017fpn}. 

For online MVS methods, feature extraction networks are usually chosen in line with the real-time operation goal. 
DeepVideoMVS~\cite{duzceker2021deepvideomvs} combines MNasNet~\cite{tan2019mnasnet}, which is lightweight and has low latency, with FPN. 
SimpleRecon~\cite{sayed2022simplerecon} and DoubleTake~\cite{sayed2024doubletake} use the first two blocks of
ResNet18~\cite{he2016deep} for the cost volume and an EfficientNet-v2~\cite{tan2021efficientnetv2} encoder for the monocular image prior,  which maintains efficiency and yields a significant improvement in depth map accuracy. 

For offline MVS methods, MVSNet~\cite{yao2018mvsnet} uses a stacked 8-layer 2D CNN to extract deep features for all images. 
Coarse-to-fine methods further extract multi-scale features for estimation on multiple scales with FPN~\cite{gu2020cascade,cheng2020deep,wang2021patchmatchnet} or multi-scale RGB images~\cite{yang2020cost,GeoMVSNet}. 
Recently, many following works pay more attention to feature extraction to improve the representation power of deep features. 
\cite{wei2021aa,mi2021generalized,ding2021transmvsnet} introduce deformable convolutions to adaptively learn receptive fields for areas with varying richness of texture. 
TransMVSNet~\cite{ding2021transmvsnet} adpots a Feature Matching Transformer for robust long-range global context aggregation within and across images. 
With FPN as the main feature extractor, WT-MVSNet~\cite{liao2022wt} and MVSFormer++~\cite{cao2024mvsformer++}, further introduce Vision Transformers~\cite{caron2021emerging,chu2021twins,liu2021swin} for image feature enhancement.

\subsection{Cost Volume Construction}\label{sec:cvc}
For both online and offline MVS, the cost volume is constructed with the plane sweep algorithm as discussed in \cref{sec:ps} with depth samples $\mathbf{D} \in \mathbb{R}^{H\times W \times D}$. 

\subsubsection{Online MVS}
To reduce computation and improve the efficiency for online applications, online MVS methods usually construct 3D cost volumes $\mathbf{C} \in \mathbb{R}^{H\times W \times D}$. The cost volume stores a single value as matching cost for each pixel $\mathbf{p}$ and depth sample $d$, which we denote as $\mathbf{C}(\mathbf{p}, d)$. 
For example, MVDepthNet~\cite{wang2018mvdepthnet} and GP-MVS~\cite{hou2019multi} compute the per-pixel intensity difference between the reference and each source view and average the matching costs from all source views:

\begin{equation}
    \mathbf{C}(\mathbf{p}, d) = \frac{1}{N-1} \sum_{i=1}^{N-1} AD(\mathbf{I}_0 (\mathbf{p}), \mathbf{I}_i(\mathbf{p}_{i}(d))),
\end{equation}
where $AD(\cdot, \cdot)$ denotes the absolute intensity difference, $\mathbf{p}_{i}(d)$ is the reprojected pixel computed from \cref{eq:point_project}. 
Instead of intensity difference, Neural RGB-D~\cite{liu2019neural} computes the $L_2$ norm of the difference between the reference and warped source features as the matching cost.
DeepVideoMVS~\cite{duzceker2021deepvideomvs} computes the dot product of the reference and warped source features:

\begin{equation}
    \mathbf{C}(\mathbf{p}, d) = -\frac{1}{C \times (N-1)} \sum_{i=1}^{N-1} \langle \mathbf{F}_0 (\mathbf{p}), \mathbf{F}_i(\mathbf{p}_{i}(d)) \rangle,
\end{equation}
where $\mathbf{F}_0, \mathbf{F}_i$ are image features, and $C$ is the number of feature channels. 
Based on DeepVideoMVS, SimpleRecon~\cite{sayed2022simplerecon} further introduces metadata, \eg, ray direction and relative pose, into the cost volume, which is found to increase the accuracy. %
DoubleTake~\cite{sayed2024doubletake} further includes depth hint, the depth map rendered from the previously predicted geometry, into the cost volume. 

\subsubsection{Offline MVS}
To encode more matching information and improve reconstruction quality, offline MVS methods~\cite{yao2018mvsnet,gu2020cascade} usually construct 4D cost volumes $\mathbf{C} \in \mathbb{R}^{H\times W \times D \times C}$, where each pixel $\mathbf{p}$ and depth sample $d$ is associated with a matching cost of dimension $C$. %
Since there may exist an arbitrary number of source views and serious occlusion~\cite{schops2017multi}, robustly aggregating matching information from all source views is an important step. 
MVSNet~\cite{yao2018mvsnet} proposes a variance-based cost volume as follows:
\begin{equation}
\label{eq:cost-volume}
\mathbf{C}={\text{Var}}\left(\mathbf{V}_{0}, \cdots, \mathbf{V}_{N-1}\right)=\frac{\sum_{i=0}^{N-1}\left(\mathbf{V}_{i}-\overline{\mathbf{V}}\right)^{2}}{N},
\end{equation}
where $\mathbf{V}_0$ is the reference feature volume, and $\{\mathbf{V}_i\}_{i=i}^{N-1}$ are warped feature volumes of source views, 
$\overline{\mathbf{V}}=1/N \sum_{i=0}^{N-1} \mathbf{V}_i$ is the average feature volume. 
Specifically, $\mathbf{V}_0 (\mathbf{p}, d)=\mathbf{F}_0(\mathbf{p})$, $\mathbf{V}_i (\mathbf{p}, d) = \mathbf{F}_i(\mathbf{p}_{i}(d))$.
To reduce computation, CIDER~\cite{xu2020learning_inverse} adopts group-wise correlation~\cite{guo2019group} to compute a lightweight cost volume between reference and warped features from $N-1$ source views. Then the individual cost volumes for different source images are averaged as the final cost volume $\mathbf{C}$. 
Without considering occlusions, these methods consider all source views equally with the averaging operation. 

Nonetheless, it is essential to emphasize that \textit{occlusions} are crucial to consider, as they pose common challenges in MVS, frequently leading to invalid matching and inaccurate estimations~\cite{schonberger2016pixelwise}. Incorporating visibility information to aggregate matching details from source views can substantially bolster robustness against occlusions, thereby enhancing the accuracy of the reconstruction process~\cite{schonberger2016pixelwise}.

To estimate view weights for source views, PVA-MVSNet~\cite{yi2020pyramid} applies gated convolution~\cite{yu2019free} to adaptively aggregate cost volumes. 
View aggregation tends to give occluded areas smaller weights and the reweighting map is yielded according to the volume itself. 
Vis-MVSNet~\cite{zhang2020visibility} aggregates pair-wise cost volumes by weighted sum, where the weight is negatively related to the uncertainty of depth probability distribution. 
~\cite{xu2022learning,wang2021patchmatchnet,wang2021itermvs} utilize pixel-wise view weight networks to learn view weights from the pair-wise cost volumes without supervision.

\subsection{Cost Volume Regularization}\label{sec:cvr}
Usually, the raw cost volume constructed from image features may be noisy and should be incorporated with smoothness constraint for depth estimation~\cite{yao2018mvsnet}. 
Therefore, cost volume regularization is an important step to refine the raw cost volume and aggregate matching information from a large receptive field. %

\subsubsection{Online MVS}
A 2D encoder-decoder architecture is commonly used to aggregate information. 
MVDepthNet~\cite{wang2018mvdepthnet} and GP-MVS~\cite{hou2019multi} concatenate the reference image and cost volume and send them to an encoder-decoder architecture with skip connections. 
In addition, GP-MVS~\cite{hou2019multi} uses Gaussian process to fuse information from previous views. 
Neural-RGBD~\cite{liu2019neural} accumulates depth probability volumes over time with a Bayesian filtering framework to effectively reduce depth uncertainty and improve robustness and temporal stability. 
DeepVideoMVS~\cite{duzceker2021deepvideomvs} applies 2D U-Net~\cite{ronneberger2015u} on the cost volume and adds skip connections between the image encoder and cost volume encoder at all resolutions. It further uses ConvLSTM~\cite{xingjian2015convolutional} to propagate past information with a small overhead of computation
time and memory consumption. 
SimpleRecon~\cite{sayed2022simplerecon} fuses multi-scale image features, extracted from the pretrained EfficientNetv2~\cite{tan2021efficientnetv2}, into the cost volume encoder to improve the performance. 

\subsubsection{Offline MVS}
Among most offline MVS methods that use 4D cost volumes, there are three main categories of cost volume regularization: direct 3D CNN, RNN, and coarse-to-fine. %

\noindent \textbf{Direct 3D CNN:}
Similar to stereo estimation~\cite{kendall2017end,psmnet,guo2019group}, 3D CNN is widely used for cost volume regularization in MVS~\cite{yao2018mvsnet,im2018dpsnet,chen2019point,luo2019pmvsnet,xu2020learning_inverse}. While this concept has been explored for stereo matching in works like DPSNet~\cite{im2018dpsnet}, MVSNet~\cite{yao2018mvsnet} serves as the blueprint for learning-based offline MVS methods. It adopts a 3D U-Net~\cite{ronneberger2015u}to regularize the cost volume, which aggregates context information from a large receptive field with relatively low computation cost. 
It is found that the 3D regularization can capture better geometry structures, perform photometric matching in 3D space, and alleviate the influence of image distortion caused by perspective transformation and occlusions~\cite{chen2019point}. 
However, since 3D CNN is memory and run-time consuming, many offline methods~\cite{yao2018mvsnet,chen2019point,luo2019pmvsnet,xu2020learning_inverse} use limited depth hypotheses and estimate depth maps at low resolution.

\noindent \textbf{RNN:}
Instead of 3D CNN, R-MVSNet~\cite{yao2019recurrent} sequentially regularizes 2D slices of the cost volume along the depth dimension with a convolutional GRU~\cite{cho2014learning}, which is able to gather spatial and uni-directional context information in the depth dimension. 
$D^2$HC-RMVSNet~\cite{yan2020dense} augments R-MVSNet~\cite{yao2019recurrent} with a complex convolutional LSTM~\cite{xingjian2015convolutional}. 
AA-RMVSNet~\cite{wei2021aa} further introduce an intra-view feature aggregation module for feature extraction and an inter-view cost volume aggregation module to adaptively aggregate cost volumes of different views. 
With sequential process of 2D slices, these methods improve the scalability for high-resolution reconstruction as well as large-scale scenes and reduce memory, however, at the cost of run-time~\cite{wei2022bidirectional}.

\noindent \textbf{Coarse-to-Fine:}
Predicting depth in a coarse-to-fine manner~\cite{gu2020cascade, cheng2020deep, yang2020cost, wang2021patchmatchnet,mi2021generalized, xu2022learning, wang2022mvster} is another solution of reducing both memory consumption and running-time. 
A coarse depth map is first predicted and then upsampled and refined during finer stages to construct fine details. %
CasMVSNet~\cite{gu2020cascade} constructs cascade cost volumes by warping features with reduced depth ranges around the previous coarse depth maps. Based on \cref{equ:homo}, the homography of differential warping at stage $k+1$ is: 
\begin{equation}
    \mathbf{H}_i^{(k+1)}(d^{(k)}+\Delta^{(k+1)}) = (d^{(k)}+\Delta^{(k+1)})\mathbf{K}_0\mathbf{T}_0\mathbf{T}^{-1}_i\mathbf{K}^{-1}_i,
\end{equation}
where $d^{(k)}$ is the estimated depth value at stage $k$ and $\Delta^{(k+1)}$ is the residual depth to be determined in the current stage. 
Following MVSNet~\cite{yao2018mvsnet}, 3D CNN is used on each stage to regularize the cost volume. 
UCS-Net~\cite{cheng2020deep} estimates the uncertainty from coarse prediction to adaptively adjust search ranges in the finer stages. 
CVP-MVSNet~\cite{yang2020cost} constructs the cost volume with a proposed optimal depth resolution of half pixel to narrow depth range in finer stages. 
EPP-MVSNet~\cite{ma2021epp} introduces an epipolar-assembling module to assemble high-resolution information into cost volume and an entropy-based process to adjust depth range. 
WT-MVSNet~\cite{liao2022wt} uses a window-based Epipolar Transformer for enhanced patch-to-patch
matching, and a window-based Cost Transformer to better aggregate global information. %

\subsection{Iterative Update}\label{sec:eff}
Diverging from conventional approaches, certain methods~\cite{wang2021patchmatchnet,lee2021patchmatch,wang2021itermvs,ma2022multiview,wang2022efficient} adopt iterative updates to gradually refine depth maps. 
By iteratively updating the depth maps based on successive iterations, these methods can progressively improve the accuracy and consistency of the reconstructed 3D scene. 
Moreover, the ability to control the number of iterations provides users with the flexibility to prioritize either computational efficiency or reconstruction quality, depending on specific application requirements.

Some methods~\cite{wang2021patchmatchnet,lee2021patchmatch} combine iterative PatchMatch~\cite{barnes2009patchmatch} with deep learning. 
PatchMatch algorithm is widely used in many traditional MVS methods~\cite{bleyer2011patchmatch,galliani2015massively,schonberger2016pixelwise,xu_2019_acmm}. 
The PatchMatch algorithm mainly consists of: random initialization, propagation of hypotheses to neighbors, and evaluation for choosing best solutions. 
After initialization, the approach iterates between propagation and evaluation until convergence. 
Traditional methods usually design fixed patterns for propagation. %
Recently, PatchmatchNet~\cite{wang2021patchmatchnet} proposes learned adaptive propagation and cost aggregation modules, which enables PatchMatch to converge faster and deliver more accurate depth maps. 
PatchMatch-RL~\cite{lee2021patchmatch}, jointly estimates depth, normal and visibility with a coarse-to-fine structure. Considering \textit{argmax} based hard decisions/sampling of PatchMatch is non-differentiable, PatMatch-RL adopts reinforcement learning in training.

Recently, RAFT~\cite{teed2020raft} estimates optical flow by iteratively updating a motion field with GRU. %
In MVS, several methods~\cite{wang2021itermvs,wang2022efficient,ma2022multiview} also adopted this approach to enhance efficiency and flexibility. 
IterMVS~\cite{wang2021itermvs} proposes a lightweight GRU-based probability estimator that encodes the per-pixel probability distribution of depth in its hidden state. %
CER-MVS~\cite{ma2022multiview} and Effi-MVS~\cite{wang2022efficient} both embed the RAFT module in a coarse-to-fine structure, which outputs a residual that is added to the previous depth. 
MaGNet~\cite{bae2022multi} iteratively updates the Gaussian distribution of depth for each pixel with the matching scores. 
DELS-MVS~\cite{sormann2023dels} proposes Epipolar Residual
Network to search for the corresponding point in the source image directly along the corresponding epipolar line and follows an iterative manner to narrow down the search space. 
IGEV-MVS~\cite{xu2023iterative} iteratively updates a disparity map regressed from geometry encoding cost volumes and pairs of correlation volumes.
Wang \etal~\cite{wang2025lightweight} introduce diffusion model in MVS and propose a conditional diffusion network that iteratively denoises the depth map, achieving high accuracy and efficiency.

\subsection{Depth Estimation}\label{sec:depth_est}
\subsubsection{Online MVS}
Many methods~\cite{wang2018mvdepthnet,hou2019multi,duzceker2021deepvideomvs, sayed2022simplerecon} apply encoder-decoder on the cost volume $\mathbf{C}$ and reduce the feature channel to 1 as $\mathbf{C}' \in \mathbb{R}^{H\times W}$. Then the \textit{sigmoid} activation $\sigma$ is applied for normalization. Together with the predefined depth range $[d_{\textrm{min}}, d_{\textrm{max}}]$ (mostly manually set), the depth map can be computed. For example, DeepVideoMVS~\cite{duzceker2021deepvideomvs} estimates depth as:
\begin{equation}
    \hat{d}(\mathbf{p}) = \left( \left(\frac{1}{d_{\textrm{min}}} - \frac{1}{d_{\textrm{max}}} \right) \cdot \sigma(\mathbf{C}'(\mathbf{p})) + \frac{1}{d_{\textrm{max}}} \right)^{-1},
\end{equation}
where $\mathbf{p}$ is the pixel coordinate. 
For other methods, Neural-RGBD~\cite{liu2019neural} directly models the cost volume as probability volume $\mathbf{P}$ and adopts \textit{soft argmax}~\cite{kendall2017end} to predict depth,~\cref{eq:softargmin}. 
MaGNet~\cite{bae2022multi} takes the mean of Gaussian distribution at last iteration as depth. 

\subsubsection{Offline MVS}
For a cost volume $\mathbf{C} \in \mathbb{R}^{H\times W\times D\times C}$, a probability volume $\mathbf{P} \in \mathbb{R}^{H\times W\times D}$ is usually generated after cost volume regularization, which is then used for depth estimation. Currently, almost all the learning-based MVS methods use exclusively either regression (\textit{soft argmax}) or classification (\textit{argmax}) to predict depth. 

Following GCNet~\cite{kendall2017end}, MVSNet~\cite{yao2018mvsnet} uses \textit{soft argmax} to regress the depth map with sub-pixel precision.  
Specifically, the expectation value along the depth direction of probability volume $\mathbf{P}$ is computed as the final prediction: 
\begin{equation}\label{eq:softargmin}
    \hat{d}(\mathbf{p}) = \sum_{i=1}^D d_i \cdot \mathbf{P}(\mathbf{p}, i),
\end{equation}
where $\mathbf{p}$ is the pixel coordinate, $d_i$ is $i$-th depth sample and $\mathbf{P}(\mathbf{p}, i)$ is the predicted depth probability. 
For coarse-to-fine methods~\cite{gu2020cascade,cheng2020deep,yang2020cost}, \textit{soft argmax} is applied on each stage to regress the depth maps. 
On the other hand, some methods~\cite{xu2020learning_inverse,wang2021patchmatchnet} compute the expectation value of \textit{inverse} depth samples since this sampling is more suitable for complex and large-scale scenes~\cite{xu2020learning_inverse}. 

In contrast, methods~\cite{yao2019recurrent,yan2020dense,wei2021aa} that use RNN for cost volume regularization mainly adopt \textit{argmax} operation. They choose the depth sample with the highest probability as the final prediction, which is similar to classification. 
Since the \textit{argmax} operation adopted by winner-take-all cannot produce depth estimations with sub-pixel accuracy, the depth map may be refined in post-processing~\cite{yao2019recurrent}.

Recently, Wang \etal~\cite{wang2021itermvs} propose a hybrid strategy to combine regression and classification. 
Similarly, Peng~\etal~\cite{peng2022rethinking} %
first use classification to get the optimal hypothesis and then regress the proximity for it.

\subsection{Depth Refinement}
Given that the raw depth estimation from MVS may be noisy, refinement is usually used to improve the accuracy. 
DeepMVS~\cite{huang2018deepmvs} apply DenseCRF~\cite{krahenbuhl2011efficient} to the coarse disparity map to encourage the pixels which are spatially close and with similar colors to have closer disparity predictions. 
R-MVSNet~\cite{yao2019recurrent} enforces multi-view photo-consistency to alleviate the stair effect and achieve sub-pixel precision.
Point-MVSNet~\cite{chen2019point} %
uses PointFlow to refine the point cloud iteratively by estimating
the residual between the depth of the current iteration
and that of the ground truth. 
Fast-MVSNet~\cite{yu2020fast} adopts an efficient Gauss-Newton layer to optimize the depth map by minimizing the feature residuals. 
PatchmatchNet~\cite{wang2021patchmatchnet} refines the final upsampled depth map with a depth residual network~\cite{hui2016depth} and reference image feature. 
~\cite{wang2021itermvs,wang2022efficient,ma2022multiview} use the mask upsampling module from RAFT~\cite{teed2020raft} to upsample and refine the depth map to full resolution by computing the weighted sum of depth values in a window based on the mask. 
Based on a coarse depth map, RayMVSNet~\cite{xi2022raymvsnet,shi2023raymvsnet++} aggregates multi-view image features with an epipolar transformer and use a 1D implicit field to estimate the SDF of the sampled points and the location of the zero-crossing point. 
GeoMVSNet~\cite{GeoMVSNet} filters the depth map by geometry enhancement in the frequency domain. 
EPNet~\cite{su2023efficient} uses a hierarchical edge-preserving residual learning module to refine multi-scale depth estimation with image context features.

\subsection{Confidence Estimation}\label{sec:confidence}
As discussed in \cref{sec:dff}, photometric confidence is important to filter out unreliable estimations during depth fusion. %
Following MVSNet~\cite{yao2018mvsnet}, most offline MVS methods take the probability of the estimation~\cite{yao2019recurrent,yan2020dense,wei2021aa} or the probability sum over several samples near the estimation~\cite{yao2018mvsnet, gu2020cascade, wang2021patchmatchnet} from the probability volume as confidence.
In stereo matching, some methods learn confidence from disparity~\cite{poggi2016learning}, RGB image~\cite{fu2017stereo,tosi2018beyond} or matching costs~\cite{kim2019laf} and obtain confidence scores in $[0,1]$ interval. 
Motivated by this, some traditional MVS methods~\cite{li2020confidence, kuhn2020deepc} based on classical PatchMatch~\cite{bleyer2011patchmatch} propose to estimate the confidence with deep learning and use this to refine the results from PatchMatch. 
Recently, IterMVS~\cite{wang2021itermvs} estimates confidence from the hidden state of a convolutional GRU. A 2D CNN followed by a \textit{sigmoid} is applied to the hidden state to predict the confidence. 
DELS-MVS~\cite{sormann2023dels} feeds pixel-wise entropy of the partition probabilities, which is computed for each source image, into a confidence network to learn confidence. The confidence is used to guide the fusion of multiple depth maps.

\subsection{Loss Function}\label{sec:loss}
\subsubsection{Online MVS}
Many methods~\cite{wang2018mvdepthnet,hou2019multi,duzceker2021deepvideomvs} compute the regression loss of estimated inverse depth maps for training: 
\begin{equation}
    L = \sum_{\mathbf{p}} \|\frac{1}{d(\mathbf{p})} -\frac{1}{\hat{d}(\mathbf{p})}\|_1,
\end{equation}
where $\mathbf{p}$ denotes the pixel coordinate, $d(\mathbf{p})$ denotes the ground truth depth, $\hat{d}(\mathbf{p})$ denotes the estimation and $||\cdot||_1$ denotes the $L_1$ loss. 
For other methods, Neural-RGBD~\cite{liu2019neural} uses Negative-Log Likelihood (NLL) over the depth with its depth probability volume. 
Similarly, MaGNet~\cite{bae2022multi} uses NLL loss since the per-pixel depth is modeled as Gaussian distribution. 
SimpleRecon~\cite{sayed2022simplerecon} computes the regression loss with log-depth. To improve performance, SimpleRecon further uses gradient loss on depth, normal loss where normal is computed with depth and intrinsics, and multi-view regression loss. 

\subsubsection{Offline MVS}
Based on the depth estimation strategy as discussed in \cref{sec:depth_est}, loss functions can be mainly categorized into regression and classification. 
For methods~\cite{yao2018mvsnet,xu2020learning_inverse} that predict depth with \textit{soft argmax}~\cite{kendall2017end}, (smooth) $L_1$ loss is usually adopted as the loss function, which is stated as:
\begin{equation}
    L = \sum_{\mathbf{p}} \|d(\mathbf{p}) -\hat{d}(\mathbf{p})\|_1.
\end{equation}
For coarse-to-fine methods~\cite{gu2020cascade,cheng2020deep,yang2020cost,wang2021patchmatchnet}, the $L_1$ loss is computed on each stage for multi-stage supervision. 

For methods\cite{yao2019recurrent,yan2020dense,wei2021aa} that predict depth with an \textit{argmax} operation, cross entropy loss is commonly used for the loss function since the problem is multi-class classification. The cross entropy loss function is defined as:
\begin{equation}
    L = \sum_{\mathbf{p}} \left( \sum_{i=1}^D -G(i, \mathbf{p}) \cdot \log[\mathbf{P}(\mathbf{p}, i)] \right),
\end{equation}
where %
$G(\mathbf{p}, i)$ is the ground truth one-hot vector of depth at pixel $\mathbf{p}$, and $\mathbf{P}(\mathbf{p}, i)$ is the predicted depth probability. 

For methods that predict depth with a hybrid strategy of classification and regression, Wang~\etal~\cite{wang2021itermvs} adopt both $L_1$ loss and cross entropy loss to supervise the regression and classification respectively, while Peng~\etal~\cite{peng2022rethinking} use focal loss~\cite{lin2017focal}. 

\section{Unsupervised \& Semi-supervised MVS with Depth Representation}\label{sec:unsup}

In this section, we introduce unsupervised and semi-supervised MVS with depth representation. 
Supervised MVS methods mentioned in \cref{sec:methods} depend extensively on the availability of accurate ground truth depth maps obtained through depth-sensing equipment. 
To make MVS practical in more general real-world scenarios, it is vital to consider alternative unsupervised methods that can provide competitive accuracy compared to the supervised ones without any ground truth. Existing unsupervised methods are built on the assumption of photometric consistency (\cref{sec:pca}), %
and are categorized into end-to-end (\cref{sec:e2e}) and multi-stage (\cref{sec:ms}). To the best of our knowledge, SGT-MVSNet~\cite{kim2021just} is the only semi-supervised method so far, and we introduce it in \cref{sec:semi}.

\subsection{Photometric Consistency Assumption}\label{sec:pca}

In the realm of unsupervised depth map prediction, extant methods~\cite{khot2019learning,xu2021digging,yang2021self,darmon2021deep} endeavor to establish photometric consistency between reference and source views. This pivotal notion revolves around the augmentation of similarity between the reference image $\mathbf{I}_{0}$ and source image $\mathbf{I}_{i}$ after warping to align with the reference view. 
Specifically, for each pixel $\mathbf{p}_0$ in $\mathbf{I}_{0}$, we use the estimated depth value $\hat{d}(\mathbf{p}_0)$ to project $\mathbf{p}_0$ into $\mathbf{I}_{i}$ with intrinsic and extrinsic matrices, following \cref{eq:point_project}. To get the warped source image $\hat{\mathbf{I}}_0^i$, we interpolate the RGB values from the source image $\mathbf{I}_i$ at the coordinates $\mathbf{p}_i$ with bilinear interpolation. Additionally, alongside the warped image $\hat{\mathbf{I}}_{0}^{i}$, a binary mask $M_{i}$ is commonly generated. This mask is employed to exclude pixels that have been projected beyond the boundaries of the image and are, thus, considered invalid.

The photometric consistency loss $ L_{\textrm{PC}}$ can be written as:
\begin{equation}
\begin{split}
 L_{\textrm{PC}}=&\sum_{i=1}^{N-1} \frac{1}{\left\|M_{i}\right\|_{1}} (\left\|(\hat{\mathbf{I}}_{0}^{i}-\mathbf{I}_{0}) \odot M_{i}\right\|_{2} \\&+ \left\|(\nabla \hat{\mathbf{I}}_{0}^{i}-\nabla \mathbf{I}_{0}) \odot M_{i}\right\|_{2}),
\end{split}
\end{equation}
where $\nabla$ denotes the gradient at the pixel level, while $\odot$ symbolizes element-wise Hadamard multiplication. 

In most cases~\cite{khot2019learning,xu2021self,xu2021digging,yang2021self,chang2022rc}, the incorporation of structural similarity loss, $L_{\textrm{SSIM}}$, and depth smoothness loss, $L_{\textrm{SM}}$, into the computation is common, which improves the stability of training process and speeds up the convergence. 
Based on the Structural Similarity Index (SSIM)~\cite{wang2004image}, structural similarity loss is computed between a synthesized image and the reference image as:

\begin{equation}
   L_{\textrm{SSIM}}=\sum_{i=1}^{N-1}\left[1-\operatorname{SSIM}\left(\mathbf{I}_{0}, \hat{\mathbf{I}}_{0}^{i}\right)\right] \odot M_{i},
\end{equation}

\begin{equation}
    \operatorname{SSIM}(x, y)=\frac{\left(2 \mu_{x} \mu_{y}+c_{1}\right)\left(2 \sigma_{x y}+c_{2}\right)}{\left(\mu_{x}^{2}+\mu_{y}^{2}+c_{1}\right)\left(\sigma_{x}^{2}+\sigma_{y}^{2}+c_{2}\right)},
\end{equation}
where $\mu$, $\sigma^{2}$ represent the mean and variance of the images, $c_1,c_2$ are constants to avoid numerical issues. %

The incorporation of a smoothness loss term serves to promote the continuity of depth information within the context of image and depth disparity alignment. This continuity is evaluated based on the color intensity gradient present in the input reference image. The smoothness loss, $L_{\textrm{SM}}$, is defined as:
\begin{equation}
L_{\textrm{SM}}= \sum_{\mathbf{x}}\left|\nabla_{u} \tilde{d}(\mathbf{x})\right| e^{-\left|\nabla_{u} \mathbf{I}_{0}(\mathbf{x})\right|}+\left|\nabla_{v} \tilde{d}(\mathbf{x})\right| e^{-\left|\nabla_{v} \mathbf{I}_{0}(\mathbf{x})\right|},
\end{equation}
where $\nabla_{u} $ and $\nabla_{v} $ refer to the gradient along x and y axis, $\tilde{d}=d / \bar{d}$ is the mean-normalized inverse depth, and $M$ represents the set of valid pixels in the reference image.

\subsection{End-to-end Unsupervised Methods}\label{sec:e2e}
End-to-end methods~\cite{khot2019learning,xu2021self,chang2022rc,zhang2022elasticmvs,xiong2023cl} train from scratch with the same input as supervised methods (\cref{sec:methods}) but without ground truth depth. They are based on photometric consistency, structural similarity, and smoothness constraints for loss terms.

Khot \etal~\cite{khot2019learning} adopts view synthesis supervision and dynamically selects the $X$ best (lowest loss) values out of $Y$ loss maps. However, photometric correspondences can be inaccurate due to non-Lambertian surfaces, camera exposure variations, and occlusions, leading to ambiguous supervision (\cref{sec:pca}).
JDACS~\cite{xu2021self} introduces semantic consistency to address these challenges, using a pre-trained network to generate semantic maps and enforcing cross-view segmentation consistency. However, this approach can struggle to converge and lacks detailed information.
RC-MVSNet~\cite{chang2022rc} enhances supervision through neural rendering by combining NeRF~\cite{mildenhall2020nerf} and cost volumes. Based on a CasMVSNet~\cite{gu2020cascade} backone that is supervised by photometric consistency, it uses rendering consistency and view synthesis loss to handle occlusions and varying lighting. 
ElasticMVS~\cite{zhang2022elasticmvs} addresses issues in photometric loss-based geometry by introducing a part-aware patch-match framework, using an elastic part representation to guide the depth map prediction. The network is optimized with contrastive and spatial concentration losses to promote pixel isolation.
CL-MVSNet~\cite{xiong2023cl} improves proximity between positive pairs through contrastive consistency between a regular CasMVSNet branch and two contrastive branches. It also introduces $L_{0.5}$ photometric consistency loss, which is more robust to outliers because it penalizes large errors less severely than traditional $L_1$ or $L_2$ norms. By growing sub-linearly with the error, it prevents the model's training from being dominated by outliers with large errors. This allows the network to focus on well-matched regions. 

These end-to-end methods train from scratch without pre-processing, reducing training time and complexity in real-world applications.

\subsection{Multi-stage Unsupervised Methods}\label{sec:ms}
Multi-stage methods require pre-training or data pre-processing and are typically based on pseudo-label generation. 
Self-supervised CVP-MVSNet~\cite{yang2021self} generates pseudo ground truth depth using photometric consistency, geometric consistency and point cloud fusion~\cite{yao2018mvsnet}. %
U-MVS~\cite{xu2021digging} pre-trains an optical-flow network, PWC-Net~\cite{sun2018pwc}, and uses flow-depth consistency to generate pseudo labels, reducing supervision ambiguity in foregrounds and backgrounds with an uncertainty-aware self-training consistency.
KD-MVS~\cite{ding2022kdmvs} employs knowledge distillation to increase performance. A teacher MVS model generates pseudo ground-truth labels through photometric and feature-metric consistency loss, with uncertainty encoding. These labels are then used to train the student models.

\subsection{Semi-supervised Methods}\label{sec:semi} 
SGT-MVSNet~\cite{kim2021just} uses sparse 3D points to estimate depth maps. A 3D point consistency loss minimizes differences between back-projected 3D points and ground truth. A coarse-to-fine depth propagation module improves accuracy at edges and boundaries.

\section{Learning-based MVS with Other Representations}\label{sec:other_mvs}
n this section, we discuss learning-based MVS with the following representations: voxel grids, NeRF, 3D Gaussian Splatting, and direct point-based representations. These approaches have demonstrated impressive 3D reconstruction capability in recent years.

\subsection{Voxel-based Methods}
These models~\cite{murez2020atlas, sun2021neuralrecon, bozic2021transformerfusion, stier2021vortx} estimate the scene geometry with volumetric representation by leveraging implicit function, \eg, SDF. 
Specifically, these methods %
predict the TSDF volume from the 3D feature volume constructed by lifting 2D image features and extract the mesh with marching cube~\cite{lorensen1998marching}.

\subsubsection{Direct 3D CNN}
\begin{figure}[tb]
    \centering
    \includegraphics[width=\linewidth]{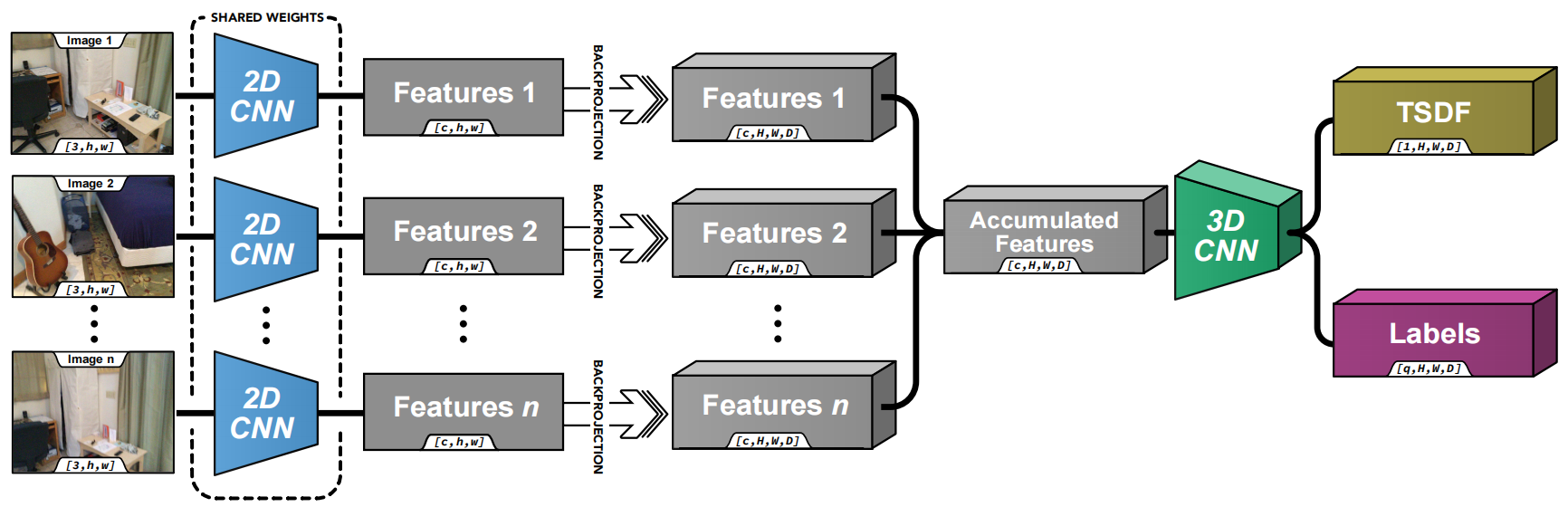}
    \caption{Pipeline of Atlas~\cite{murez2020atlas}. 2D image features are back-projected into 3D volumes, which are aggregated and passed through a 3D CNN to directly regress a TSDF volume. } 
    \label{fig:atlas}
\end{figure}

As a seminal work, Atlas~\cite{murez2020atlas} employs a 3D CNN with four
scale resolution pyramid to regress the TSDF volume from the feature volume accumulated from all images, as shown in \cref{fig:atlas}. 
The TSDF reconstructions are supervised using $L_1$ loss to the ground truth TSDF vaules. 
However, constructing a dense 3D feature volume is a huge computational overhead for large scenes.

\subsubsection{Coarse-to-fine}
It is a common practice to adopt a coarse-to-fine paradigm~\cite{sun2021neuralrecon,bozic2021transformerfusion,stier2021vortx} for the sake of efficiency. One of its core steps is voxel sparsification, which eliminates empty voxels at the coarse level so that memory consumption is reduced in the fine level. 
NeuralRecon~\cite{sun2021neuralrecon} incrementally reconstructs the scene in a coarse-to-fine and fragment-wise manner. 
First, in order to reduce memory, NeuralRecon predicts TSDF with a three-level coarse-to-fine approach that gradually increases the density of sparse voxels. 
Second, it reconstructs the TSDF volume of a local fragment and then fuses it with the global volume by GRU~\cite{cho2014properties} Fusion. 

Based on the coarse-to-fine framework of NeuralRecon, TransformerFusion~\cite{bozic2021transformerfusion} introduces transformer~\cite{vaswani2017attention} that
enables the network to learn to attend to the most relevant image frames for each
3D location in the scene. 
Similarly, VoRTX~\cite{stier2021vortx} uses a transformer to fuse image features at each voxel to produce the multi-view feature volume.

Although state-of-the-art voxel-based methods use coarse-to-fine framework to explicitly improve the efficiency, they are still limited to indoor scenes that are strictly bounded,~\eg, ScanNet~\cite{dai2017scannet}, since the memory storage explicitly increases when the scale of scene increases,~\eg, unbounded outdoor scenes.

\subsection{NeRF-based Methods}
\begin{figure}[tb]
    \centering
    \includegraphics[width=\linewidth]{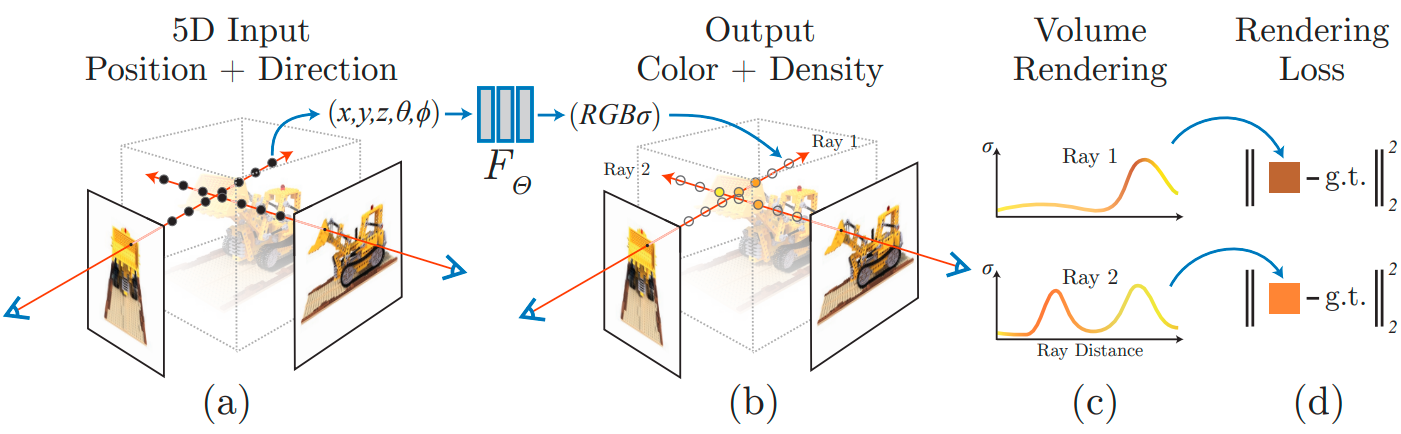}
    \caption{Pipeline of NeRF~\cite{mildenhall2020nerf}. Given a 3D position and 2D viewing direction (a), an MLP produces the color and volume density (b). Then volume rendering is used to composite these values into an image (c). The optimization is minimizing the rendering loss (d). }
    \label{fig:nerf}
\end{figure}

In novel view synthesis, Neural Radiance Field (NeRF)~\cite{mildenhall2020nerf} has kicked off a new emerging representation of 3D, which offers a differentiable volume-rendering scheme to supervise a 3D radiance-based representation with 2D image-level losses. 
NeRF employs multi-layer perceptron (MLP) to map a position $(x,y,z)$ and the normalized view direction $(\theta,\phi)$ to the corresponding color $\mathbf{c}$ and volume density $\sigma$. 
For a specific ray at a novel viewpoint, NeRF uses approximated numerical volume rendering to compute the accumulated color as: 
\begin{equation}
    \mathbf{C}=\sum_{i=1}^{N} T_i (1- \exp (-\sigma_i\delta_i))\mathbf{c}_i, 
\end{equation}
where $i$ is the index of sample, $T_i = \exp (-\sum_{j=1}^{i-1} \sigma_j\delta_j)$ is the accumulated transmittance, and $\delta_i=t_{i+1}-t_i$ is the distance between adjacent samples. The model is trained by minimizing the loss between the predicted and ground truth color:
\begin{equation}\label{eq:nerf_loss}
    \mathcal{L}_{\text{color}} = \mathbb{E} [||\mathbf{C} - \mathbf{C}_{gt} ||_2^2].
\end{equation}
The pipeline of NeRF is shown in~\cref{fig:nerf}. 
Many subsequent endeavors~\cite{sun2022direct,chen2022tensorf,muller2022instant,barron2022mip,fridovich2022plenoxels,chen2023mobilenerf,barron2023zip,reiser2023merf,gao2023strivec} further improve NeRF in quality, fast training, memory efficiency and real-time rendering.

\begin{figure*}[tb]
    \centering
    \includegraphics[width=\linewidth]{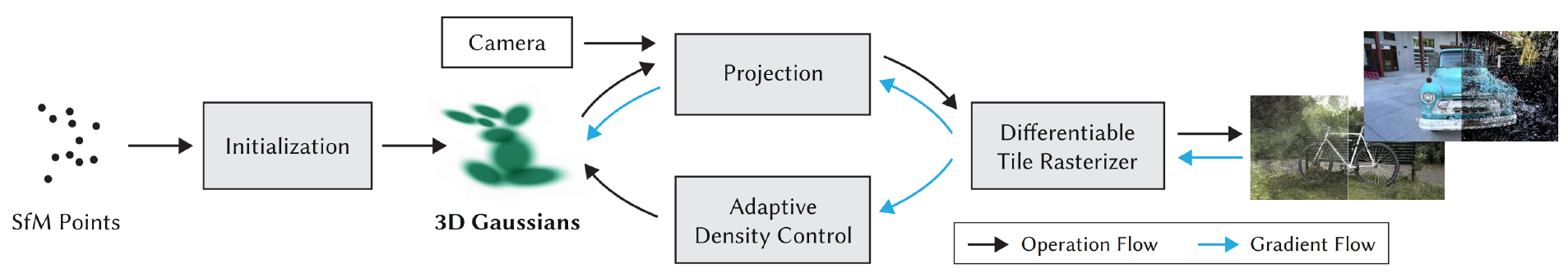}
    \caption{Pipeline of 3DGS~\cite{kerbl3Dgaussians}. 3DGS initializes 3D Gaussian primitives from scene geometry priors (e.g., SfM points) and computes pixel colors by compositing their projected attributes via 3D-to-2D splatting. }
    \label{fig:3dgs}
\end{figure*}

\begin{figure*}[tb]
    \centering
    \includegraphics[width=\linewidth]{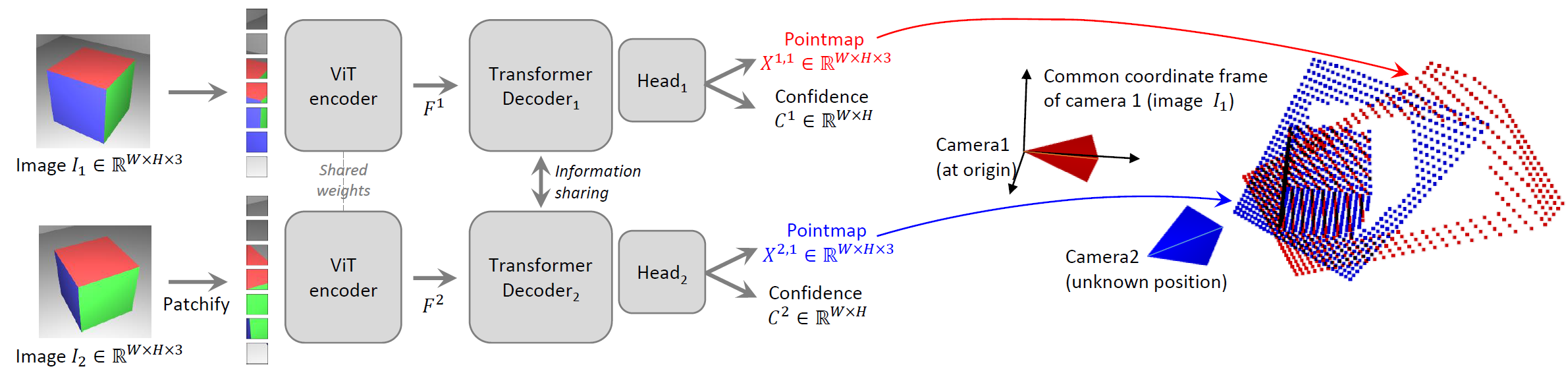}
    \caption{Pipeline of DUSt3R~\cite{wang2023dust3r}. Two views of a scene $(I^1, I^2)$ are first encoded in a Siamese manner with a shared ViT encoder.
The resulting token representations $F^1$ and $F^2$ are then passed to two transformer decoders that constantly exchange information via cross-attention. Finally, two regression heads output the two corresponding pointmaps and associated confidence maps. Importantly, the two pointmaps are expressed in the same coordinate frame of the first image $I^1$. }
    \label{fig:dust3r}
\end{figure*}

Though the initial purpose of NeRF is to perform novel view synthesis, VolSDF~\cite{yariv2021volume} and NeuS~\cite{wang2021neus} integrate NeRF with SDF for surface reconstruction. The SDF, denoted as $f$, is transformed into the density $\sigma$ for volume rendering.  
After training, the mesh can be extracted from the SDF field with Marching Cubes~\cite{lorensen1998marching}. Since the main focus of this survey is 3D reconstruction, we discuss these NeRF-based methods with SDF representation in the following.

\subsubsection{Optimization-based NeRF}
Without ground truth geometry supervision, \eg, depth map or TSDF volume, training is mainly done in a self-supervised manner with rendering loss, \cref{eq:nerf_loss}. To enforce SDF function $f$ approximate a valid SDF, existing methods~\cite{yariv2021volume,wang2021neus} use an eikonal loss~\cite{gropp2020implicit} for regularization:

\begin{equation}
    \mathcal{L}_{\text{eik}} = \mathbb{E}_{\mathbf{x}} [(||\nabla f(\mathbf{x}) ||_2 - 1 )^2],
\end{equation}
where $\mathbf{x}$ denotes the 3D point. Motivated by Instant-NGP~\cite{muller2022instant} that accelerates training with hash grids, many methods~\cite{wang2023neus2, li2023neuralangelo, rosu2023permutosdf, wang2023unisdf} use hash grids to speed up training and improve surface details. 

However, these methods still mainly use pixel-wise rendering loss for training, which may result in geometric ambiguity, especially in textureless areas~\cite{wei2021nerfingmvs}. 
To alleviate this issue, some methods~\cite{darmon2022improving,fu2022geo} use patch-wise loss to reduce ambiguity. Specifically, based on rendered depth and normal, they warp reference image patches to neighboring views and measure photometric consistency, which is similar to unsupervised depth map-based MVS. 
In addition, other methods introduce explicit geometry supervision,~\eg, monocular depth/normal priors~\cite{yu2022monosdf,wang2022neuris} and sparse SfM point cloud~\cite{fu2022geo}. For example, MonoSDF~\cite{yu2022monosdf} uses a pre-trained Omnidata model~\cite{eftekhar2021omnidata} to predict monocular depth and normal, which are used to enforce consistency with rendered depth and normal. 
Recently, Ref-NeRF~\cite{verbin2022ref}, a variant of original NeRF, reparameterizes the appearance with separate diffuse and reflective components by using the reflected view direction, which improves the rendering of specular surfaces. Therefore, some follow-up methods~\cite{yariv2023bakedsdf,liu2023nero,liang2023envidr,ge2023ref,wang2023unisdf} adopt this representation in reconstruction and can successfully reconstruct specular surfaces. 

Recall that the depth map-based MVS methods predict depth with photometric consistency across multiple views. However, this assumption fails for glossy surfaces, and thus these methods cannot reconstruct them accurately. Therefore, we believe that NeRF-based MVS is more suitable than depth map-based MVS for reconstructing specular surfaces.

\subsubsection{Generalizable NeRF}
NeRF usually requires a lengthy optimization for each scene and cannot generalize to unseen scenes in a zero-shot manner.
To address this issue, recent methods~\cite{chen2021mvsnerf, long2022sparseneus,ren2022volrecon,liang2023retr} propose generalizable NeRF pipelines for both rendering and reconstruction.
Interestingly, we observe that the majority of these methods are inspired by depth map-based and voxel-based MVS.
For example, MVSNeRF~\cite{chen2021mvsnerf} and C2F2NeUS~\cite{xu2023c2f2neus} construct cost volumes with plane sweep to encode global features, following depth map-based offline MVS methods~\cite{yao2018mvsnet,gu2020cascade}.
Similar to voxel-based MVS~\cite{murez2020atlas}, some methods~\cite{long2022sparseneus,ren2022volrecon,liang2023retr} use image features to construct feature volumes for geometry estimation.
More recently, large reconstruction models (LRMs)~\cite{hong2023lrm} leverage transformer backbones to directly predict NeRF representations from single or sparse images in a feed-forward manner. These models support inputs ranging from single image~\cite{hong2023lrm,han2024vfusion3d,tochilkin2024triposr} to posed~\cite{xu2023dmv3d,LaRa} and un-posed multi-view imagery~\cite{wang2023pf}, significantly improving practicality beyond per-scene optimization.

\subsection{3D Gaussian Splatting-based Methods}

Unlike implicit representations with a coordinate-based MLP such as NeRF~\cite{mildenhall2020nerf}, 3D Gaussian Splatting~\cite{kerbl20233d} (3DGS) explicity represents the scene with point primitives, each of which is parameterized as a scaled Gaussian with 3D covariance matrix $\mathbf{\Sigma}$ and mean $\mathbf{\mu}$:
\begin{equation}
    G(\mathbf{x}) = e^{-\frac{1}{2}(\mathbf{x}-\mathbf{\mu})^T \mathbf{\Sigma}^{-1} (\mathbf{x}-\mathbf{\mu})},
\end{equation}
where $\mathbf{x}$ is an arbitrary position. $\mathbf{\Sigma}$ is formulated with a scaling matrix $\mathbf{S}$ and rotation matrix $\mathbf{R}$ as:
\begin{equation}
    \mathbf{\Sigma} = \mathbf{R} \mathbf{S} \mathbf{S}^T \mathbf{R}^T.
\end{equation}
In addition, each Gaussian contains the color $\mathbf{c}$ modeled by Spherical Harmonics and an opacity $\alpha$. 
Different from NeRF that uses volume rendering, 3DGS efficiently renders the scene via tile-based rasterization. After projecting 3D Gaussian $G(\mathbf{x})$ into the 2D Gaussian $G'(x)$ on the image plane~\cite{kerbl20233d}, a tile-based rasterizer efficiently sorts the 2D Gaussians and employs $\alpha$-blending for rendering:
\begin{equation}
    \mathbf{C}(x) = \sum_{i \in N} \mathbf{c}_i \sigma_i \prod_{j=1}^{i-1}(1-\sigma_j), \quad \sigma_i = \alpha_i G'(x), 
\end{equation}
where $x$ is the pixel location, $N$ is the number of sorted 2D Gaussians. 

During training, 3DGS minimizes the rendering loss like NeRF, as in \cref{eq:nerf_loss}. 3DGS can be initialized with SfM~\cite{snavely2006photo,schonberger2016structure} or MVS~\cite{schonberger2016pixelwise}, which performs better than random initialization~\cite{kerbl20233d}. The pipeline of 3DGS is shown in~\cref{fig:3dgs}.

\subsubsection{Optimization-based 3DGS}
Motivated by NeRF-based MVS methods, researchers try to adapt 3DGS for reconstruction task. 
Though 3DGS achieves high-quality novel-view synthesis, it is challenging to recover high quality geometry since no explicit geometry constraint is used and 3D Gaussians do not correspond well to the actual surface because of the 3D covariance~\cite{guedon2023sugar}.

In order to make 3D Gaussians correspond well to the actual surface, SuGaR~\cite{guedon2023sugar} introduces a geometry regularization term that encourages the 3D Gaussians to be well-aligned over the scene surfaces so that the Gaussians can contribute to better geometry. To reconstruct the mesh, SuGaR samples 3D points on a level set of the density computed from 3D Gaussians and then runs Poisson Reconstruction~\cite{kazhdan2006poisson} on these points. 
Instead of performing regularization on 3D Gaussian, 2DGS~\cite{huang20242d} introduces a 2D Gaussian representation by modeling the 3D primitives as surfels.

Similar to NeRF, relying solely on image rendering loss can easily fall into local minima, leading to Gaussian shapes inconsistent with the actual surface. Introducing explicit geometry constraints can alleviate this problem. For example, PGSR~\cite{chen2024pgsr} enforces the local consistency of depth and normal in a local image patch, and multi-view geometric and photometric consistency as COLMAP~\cite{schonberger2016structure}. 

\subsubsection{Generalizable 3DGS}
Following the paradigm of feed-forward NeRF, PixelSplat~\cite{charatan2024pixelsplat} predicts 3DGS representations from paired images by leveraging epipolar constraints to resolve scale ambiguity. MVSplat~\cite{chen2024mvsplat} further incorporates depth cues for 3DGS prediction by constructing a cost volume from multiple input images. DepthSplat~\cite{xu2025depthsplat} extends this idea by achieving more robust depth estimation through the combination of cost volume and monocular depth features~\cite{yang2024depthv2}. 
Instead of focusing on improving geometry and rendering quality, recent works such as SplatterImage~\cite{szymanowicz2024splatter}, Flash3D~\cite{szymanowicz2025flash3d}, GS-LRM~\cite{zhang2024gs},LGM~\cite{tang2024lgm}, and Bolt3D~\cite{szymanowicz2025bolt3d} explore the generalization capability of single-image-to-3DGS generation without per-scene optimization. These methods leverage multi-view diffusion models or transformers pretrained on large-scale multi-view datasets to directly predict per-pixel 3D Gaussian attributes from images.

\subsection{Large Feed-forward Point-based Methods}
In 3D reconstruction, a prominent trend is to \emph{directly learn} 3D point geometry from large-scale multi-view data with transformer backbones~\cite{vaswani2017attention}, shifting the paradigm from per-scene optimization toward one-shot inference of geometry or renderable content. Unlike traditional MVS that relies on hand-crafted cost volumes, these methods shift the paradigm toward one-shot inference of dense point maps.

Recent feed-forward point-based methods compress the classical MVS pipeline into a single forward pass: (1) feature extraction; (2) cross-view aggregation via attention in lieu of an explicit plane-sweep cost volume; and (3) prediction heads that jointly regress cameras and dense geometry (depth/point maps). We illustrate the pipeline of DUSt3R~\cite{wang2023dust3r}, a pioneering feed-forward method, in \cref{fig:dust3r}. Below we describe how modern feed-forward methods instantiate each stage in a single forward pass. 

\noindent\textbf{ViT feature extraction.} A pre-trained transformer encoder extracts view-wise, geometry-aware features/tokens from few to many images, supporting downstream camera, depth, and pointmap prediction~\cite{wang2025vggt,wang2023dust3r,leroy2024grounding,keetha2025mapanything}. In practice, DINO models are often chosen as the ViT backbone because its self-supervised features transfer robustly across tasks and domains and exhibit emergent semantics useful for dense correspondences~\cite{oquab2023dinov2}. 

\noindent\textbf{Cross-view aggregation via attention.} Instead of building explicit plane-sweep cost volumes, global or alternating attention patterns interleave per-frame and multi-frame aggregation, allowing long-range correspondence reasoning while remaining feed-forward~\cite{wang2025vggt,wang2023dust3r}. The attention mechanism functions as a learnable and unconstrained feature matching in multi-view feed-forward 3D reconstruction. With no pose prior and unknown epipolar lines, the theoretical area to be searched is actually the whole image and the Transformer modules learn implicit correspondences in a data-driven manner, as is revealed in \cite{stary2025understanding}. While the more classical cost volumes are built with known poses and epipolar lines, where the area to be searched become a 1D segment. 

\noindent\textbf{Prediction heads for geometry and pose.} Rather than deriving geometry purely from pairwise matches, modern heads directly regress intrinsics/extrinsics together with dense depth or point maps. Early work showed that regressing point maps is effective even from two views and unifies monocular/binocular settings~\cite{wang2023dust3r}; recent multi-view extensions output cameras, depth, point maps, and tracks in one pass~\cite{wang2025vggt}. Emerging designs relax reference-view assumptions via permutation-equivariant architectures and scale to long sequences with spatial memory or causal/streaming transformers~\cite{wang2025pi,wang20243d,lan2025stream3r}.

For future work, our analysis points to two avenues for advancing feed-forward 3D reconstruction. First, to address the fundamental out-of-distribution problem, future research may focus on developing memory-efficient training paradigms that allow large-scale, many-view scenes to be incorporated directly during training. By aligning the training distribution more closely with real-world test scenarios, such approaches could substantially improve generalization to complex environments. Second, to overcome the accuracy limitations of single-pass prediction, a promising direction is to create hybrid architectures that merge feed-forward speed with iterative refinement.

\section{Discussions}\label{sec:discussion}
In this section, we summarize and discuss the performance and efficiency of learning-based MVS methods. 
Moreover, we discuss potential directions for future research. 

\begin{table*}[tb]
    \caption{Quantitative results of point cloud evaluation on Tanks \& Temples~\cite{knapitsch2017tanks}. F-score is listed (the higher, the better). MapAnything~\cite{keetha2025mapanything} reconstructs point clouds directly from unposed images. All images are used for MapAnything, except for Courthouse (50\% subset due to memory limitations). }
    \centering
    \resizebox{\textwidth}{!}{ 
    \begin{tabular}{c|c|cccccc|c}
    \hline
    Representation & Methods & Barn & Caterpillar & Courthouse & Ignatius & Meetingroom & Truck & Mean \\
    \hline
    \multirow{8}{*}{Depth map} 
    & DeepVideoMVS~\cite{duzceker2021deepvideomvs} & - & - & - & 0.01 & - & 0.01 & - \\
    & CasMVSNet~\cite{gu2020cascade} & 0.52 & 0.42 & 0.36 & 0.71 &  0.35 & 0.59 & 0.49\\ 
    & $D^2$HC-RMVSNet~\cite{yan2020dense} & 0.64 & 0.57 & \textbf{0.42} & 0.74 & 0.33 & 0.70 & 0.57\\
    & AA-RMVSNet~\cite{wei2021aa} &  0.61 & 0.58 & 0.40 & 0.77 & 0.36 &  \textbf{0.71} & 0.57\\
    & IterMVS~\cite{wang2021itermvs} & 0.62 & 0.49 & 0.38 & 0.78 & 0.38 & 0.68 & 0.55 \\
    & TransMVSNet~\cite{ding2021transmvsnet} &  0.57 & 0.47 & 0.38 &  0.73 & 0.42 & 0.63 & 0.53\\
    & MVSFormer++~\cite{cao2024mvsformer++} & 0.66 & \textbf{0.59} & \textbf{0.42} & 0.78 & 0.40 & 0.69 & 0.59\\
    & CasDiffMVS~\cite{wang2025lightweight} & 0.69 & 0.54 & \textbf{0.42} & 0.81 & \textbf{0.45} & 0.70 & \textbf{0.60} \\
    \hline
    \multirow{2}{*}{Voxel} 
    & NeuralRecon~\cite{sun2021neuralrecon} & - & 0.01 & - & 0.07 & - & - & -\\
    & VisFusion~\cite{gao2023visfusion} & - & 0.09 & - & 0.09 & 0.03 & 0.15 & -\\
    \hline
    \multirow{3}{*}{NeRF} 
    & NeuS~\cite{wang2021neus} & 0.29 & 0.29 & 0.17 & 0.83 & 0.24 & 0.45 & 0.38 \\
    & Geo-NeuS~\cite{fu2022geo} & 0.33 & 0.26 & 0.12 & 0.72 & 0.20 & 0.45 & 0.35 \\
    & Neuralangelo~\cite{li2023neuralangelo} & \textbf{0.70} & 0.36 & 0.28 & \textbf{0.89} & 0.32 & 0.48 & 0.50 \\
    \hline
    \multirow{4}{*}{3DGS} 
    & SuGaR~\cite{guedon2023sugar} & 0.14 & 0.16 & 0.08 & 0.33 & 0.15 & 0.26 & 0.19 \\
    & 2DGS~\cite{huang20242d} & 0.36 & 0.23 & 0.13 & 0.44 & 0.16 & 0.26 & 0.30 \\
    & GOF~\cite{yu2024gaussian} & 0.51 & 0.41 & 0.28 & 0.68 & 0.28 & 0.58 & 0.46 \\
    & PGSR~\cite{chen2024pgsr} & 0.66 & 0.41 & 0.21 & 0.80 & 0.29 & 0.60 & 0.50 \\
    \hline
    \multirow{1}{*}{Feed-forward Point-based}
    & MapAnything~\cite{keetha2025mapanything} & 0.14 & 0.34 & 0.10 & 0.18 & 0.12 & 0.23 & 0.19 \\ 
    \hline
    \end{tabular}
    }
    \label{tab:tank_training}
\end{table*}

\subsection{Performance Comparison}
To provide a thorough comparison across different methods, we evaluate state-of-the-art methods on Tanks \& Temples~\cite{knapitsch2017tanks} and summarize the results in~\cref{tab:tank_training}. 

For depth map-based methods, we find that online methods~\cite{duzceker2021deepvideomvs} perform worse than offline methods~\cite{gu2020cascade,ding2021transmvsnet,wang2025lightweight,cao2024mvsformer++}. 
Online methods typically sample sparse depth hypotheses from a pre-defined depth range to construct a \textit{lightweight} cost volume, which limits the accuracy in large-scale scenes. In addition, online methods are limited by causal/temporal processing and cannot exploit geometric constraints from future frames as offline methods. 

For voxel-based methods, we observe that they fail on both outdoor and indoor scenes. From the representation's perspective, we can attribute the performance gap to the following two aspects. 
(a) Voxel-based methods usually specify the spatial location of the potential voxels given the camera trajectory, with a metric voxel resolution. The inherent memory complexity and limited resolution of voxel grids are fundamental bottlenecks for large-scale, unbounded scenes. 
(b) Unlike depth-based representations, which estimate geometry primarily based on photometric consistency on the 2D image plane and allow for sharp discontinuities (suitable for complex occlusions in outdoor scenes), voxel-based SDFs impose a global 3D continuity constraint. While this constraint helps regularize geometry in texture-less indoor regions, it acts as a limiting factor in large-scale scenes by enforcing smoothness where sharp edges or fine details are expected.

For NeRF/3DGS-based methods, we observe that they achieve high accuracy in small scenes,~\eg, Ignatius, producing high-quality reconstruction with fine-grained details. However, the performance is notably worse than offline methods on large scenes,~\eg, Courthouse. Scaling NeRF/3DGS to large-scale scenes is still an open topic in research. 

For large feed-forward point-based methods, although the reconstructed point clouds appear reasonable, their quantitative accuracy is typically low. This is due to several challenges: (1) they must estimate camera poses, significantly increasing task complexity; however, models like DUSt3R skip explicit camera projection, and VGGT predicts camera tokens in parallel, bypassing camera geometry during the forward pass. This flexibility in handling unposed data, while advantageous, means that metric evaluation is highly sensitive to pose alignment and can limit accuracy. (2) These models are typically not trained on scenes with hundreds of views, creating an out-of-distribution problem for larger datasets; and (3) huge memory consumption compels the use of low-resolution images or fewer input frames, directly limiting the fidelity of the final reconstruction. 

\subsection{Efficiency Comparison} \label{sec:memory_time}
Low run-time and memory consumption are crucial in most industrial applications with limited computational power and storage, \eg, robotics and AR/VR. 

\begin{figure}[t]
    \centering
    \includegraphics[width=\linewidth]{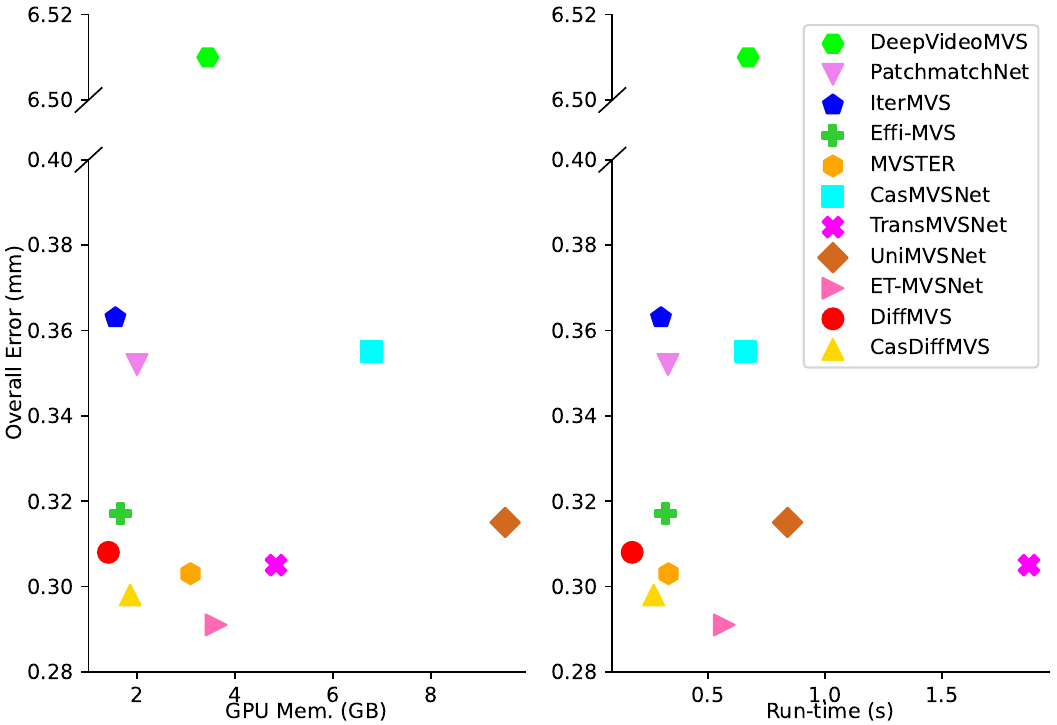}
    \caption{GPU memory consumption, run-time, and reconstruction error of state-of-the-art depth map-based methods~\cite{duzceker2021deepvideomvs,wang2021patchmatchnet,wang2022efficient,wang2021itermvs,wang2022mvster,ding2021transmvsnet,peng2022rethinking,Liu_2023_ICCV} on DTU~\cite{aanaes2016large}. To fairly compare efficiency, we fix the image resolution and the number of input images, which are set to $1600 \times 1152$ and 5 respectively. All experiments are done on the same workstation with a NVIDIA 2080 Ti GPU. }
    \vspace{-0.5cm}
    \label{fig:efficiency}
\end{figure}

Compared to other representations, depth-map based MVS is usually more suitable for these applications. Volume-based methods are difficult to scale to large scenes due to substantial memory storage. NeRF-based and 3D Gaussian Splatting-based methods typically require per-scene optimization for high-quality reconstruction. Large feed-forward methods have huge backbones and thus their GPU memory consumption is high. 
We compare the efficiency of state-of-the-art depth-map based MVS methods~\cite{gu2020cascade,wang2021patchmatchnet,duzceker2021deepvideomvs,wang2022efficient,wang2021itermvs,wang2022mvster,ding2021transmvsnet,peng2022rethinking,GeoMVSNet,Liu_2023_ICCV} in \cref{fig:efficiency}. 
Although online methods~\cite{duzceker2021deepvideomvs} achieve high efficiency in run-time with low-resolution images, we observe that their run-time and GPU memory consumption are higher than many offline methods~\cite{wang2021patchmatchnet,wang2022efficient,wang2021itermvs,wang2022mvster,wang2025lightweight} with high-resolution images. 
To improve efficiency, offline methods~\cite{wang2021patchmatchnet,wang2022efficient,wang2021itermvs,wang2022mvster,wang2025lightweight} carefully simplify the architecture,~\eg, replacing costly 3D convolution with 2D convolution. Moreover, these methods, especially CasDiffMVS~\cite{wang2025lightweight}, achieve competitive performance when compared to the top-performing methods~\cite{Liu_2023_ICCV,cao2024mvsformer++}. 
Therefore, it is promising to further improve both the performance and efficiency for practical applications.

\subsection{Future Research Directions}\label{sec:future}

\subsubsection{Datasets and Benchmarks}
For MVS methods, ScanNet~\cite{dai2017scannet} and DTU~\cite{aanaes2016large} are two main training datasets, while ScanNet~\cite{dai2017scannet}, DTU~\cite{aanaes2016large}, Tanks \& Temples~\cite{knapitsch2017tanks} and ETH3D~\cite{schops2017multi} are the main evaluation benchmarks. 

For training, the scene scale of ScanNet and DTU is relatively small (room-scale for ScanNet and object-scale for DTU) and their quality is not satisfactory~\cite{luo2020attention}. %
Recently, many researchers have tried to improve the dataset quality. 
For example, TartanAir dataset~\cite{wang2020tartanair} is a large-scale synthetic dataset that is collected in photo-realistic simulation environments. %
BlendedMVS~\cite{yao2020blendedmvs} introduces more large-scale scenes and improves the performance of many MVS methods on real-world scenes. 
MVImgNet~\cite{yu2023mvimgnet} and Objaverse~\cite{deitke2023objaverse} are large datasets of multi-view images collected by shooting videos of real-world objects or rendering 3D object models. %
However, both of them focus on object-level scenes, whereas large-scale scenes are essential for many applications. 
DL3DV~\cite{ling2024dl3dv} attempts to address this issue by capturing videos of large-scale real-world scenes. However, it lacks the ground truth geometry required for MVS training. Therefore, improving the scalability and quality of datasets is an important research direction. One direction is to use advanced sensors,~\eg, depth camera and LiDAR, to capture high-quality data in diverse real-world scenario~\cite{yeshwanth2023scannet++, wu2025indoor}. The other direction is to exploit large-scale video data,~\eg, Youtube videos, and extract high-quality pseudo ground-truth with state-of-the-art models. 

For evaluation, the main benchmarks~\cite{dai2017scannet,aanaes2016large,knapitsch2017tanks,schops2017multi} are all introduced before 2018. Some of these benchmarks are currently saturated and the methods are difficult to further improve the performance on them. 
Therefore, it is meaningful to introduce new benchmarks to evaluate the robustness and performance of learning-based methods in challenging real-world scenes. For example, the new benchmarks are expected to include challenging settings, such as dynamics, low-texture, reflections, and appearance changes~\cite{sarlin2022lamar}. %

\subsubsection{View Selection}
As discussed in \cref{sec:viewselect}, view selection is crucial for triangulation quality. Picking neighboring views that are suitable for triangulation can not only improve the reconstruction accuracy but also reduce useless computation for the bad views, \eg, views with strong occlusions. 
However, view selection is often overlooked and not well studied. For example, all offline depth map-based MVS methods~\cite{gu2020cascade,wang2021patchmatchnet,ding2021transmvsnet,wang2022mvster} follow MVSNet~\cite{yao2018mvsnet} and use the same simple heuristic strategy to compute scores for neighboring views and sort them. 
Online depth map-based MVS methods~\cite{hou2019multi, duzceker2021deepvideomvs, sayed2022simplerecon} and voxel-based methods~\cite{sun2021neuralrecon} adopt heuristic strategies to choose views with enough pose-distance. 
Though it is intractable to incorporate non-differentiable view selection into deep learning, it is worth exploring new view selection strategies since it may improve the reconstruction without changing the model design. 
Current view selection strategies simply pick the same set of neighboring views for all the pixels in the reference image, and it is probable that different reference pixels have different optimal choices of neighboring views. 
The learned pixel-wise view weight~\cite{xu2022learning,wang2021patchmatchnet,wang2021itermvs} addresses this issue by assigning different weights to source views for each pixel based on their visibility across neighboring views. %
For large feed-forward methods such as VGGT~\cite{wang2025vggt}, cross-view attention provides an alternative mechanism for adaptive view selection with the attention weights. %

\subsubsection{Feature Extraction}
Currently, most of the depth map-based~\cite{yao2018mvsnet,duzceker2021deepvideomvs} and voxel-based~\cite{sun2021neuralrecon} methods mainly use 2D CNN and FPN~\cite{lin2017fpn} structure to learn multi-scale features. To flexibly improve the receptive fields, Deformation Convolution~\cite{dai2017deformable} and attention mechanism~\cite{vaswani2017attention} are applied ~\cite{yi2020pyramid,mi2021generalized,wei2021aa,wang2021patchmatchnet,ding2021transmvsnet}. 
However, they still cannot generalize well in challenging regions, such as reflective and low-textured areas. Motivated by the success of vision transformer~\cite{dosovitskiy2020image}, recent works~\cite{cao2022mvsformer,cao2024mvsformer++,wang2023dust3r,wang2025vggt} introduce pre-trained vision transformers,~\eg, ViT~\cite{dosovitskiy2020image} and DINO~\cite{caron2021emerging,oquab2023dinov2,simeoni2025dinov3}, to capture global information, alleviating the issues of reflections and texture-less regions. 
However, vision transformers are usually trained and tested on low-resolution images due to their complexity. To accurately reconstruct the scene with high-resolution inputs, it is important to utilize fine-grained information for precise feature matching across views, which can be nicely captured with CNN. Therefore, an interesting direction is to combine vision transformer and CNN in a coarse-to-fine framework, where vision transformer is used in low-resolution and CNN is used in high-resolution for local refinement.

\subsubsection{Generative Reconstruction}
MVS tends to be sensitive to occlusions and sparse viewpoints because of their strict geometric requirements. However, these challenging conditions are quite common in practice, such as autonomous driving. 
To this end, while mainstream MVS methods follow a regression-based paradigm, it would be promising to leverage the inpainting power of generative models to produce reasonable reconstruction in unseen regions. 
For example, ReconFusion~\cite{wu2024reconfusion} explores to distill the generated novel views to a NeRF representation for sparse-view reconstruction. 
Difix3D+~\cite{wu2025difix3d+} refines noisy 2D renderings from Gaussian Splatting with reference views and Stable Diffusion~\cite{rombach2022high}. 
Achieving visually pleasing 3D reconstruction while maintaining the reconstruction fidelity and geometric consistency with generative models is a significant research trend, which will also power the development of world models.

\subsubsection{Efficiency}
Though reconstruction accuracy is considered the most important factor in MVS, it is also critical to have models that can run in near real-time and with low memory, \eg, AR/VR and autonomous driving. 
Comparatively, depth map-based methods are most suitable for achieving efficiency because of their conciseness, flexibility, and scalability. 
Though online MVS methods can achieve real-time estimation with images of low resolutions, they may have issues with memory consumption since many large backbones are usually used for feature extraction~\cite{duzceker2021deepvideomvs, sayed2022simplerecon}. In addition, the efficiency in both run-time and memory will drop when image resolution increases, as shown in~\cref{fig:efficiency}. 
For offline MVS methods, as discussed in~\cref{sec:memory_time}, some recent methods try to improve efficiency with carefully simplified network architectures. However, this usually limits the performance when compared to state-of-the-art methods. In addition to designing lightweight modules~\cite{wang2025lightweight}, using model compression techniques and knowledge distillation are potential directions to not only keep the good performance but also improve efficiency. 

Large feed-forward models heavily rely on attention mechanism to capture intra-frame and inter-frame relations. However, the high computational cost restricts the use of high-resolution or numerous input images, which can degrade performance~\cite{wang2025vggt,keetha2025mapanything}. %
To reduce computational cost of attention mechanism, one direction is to introduce token merging to reduce attention redundancy~\cite{shen2025fastvggt}.

\subsubsection{Prior Assistance}

MVS mainly relies on assessing local photometric consistency to find the optimal matching across images. Accordingly, it usually encounters difficulties when estimating the geometry for regions where the photometric measurement becomes unreliable or invalid, \eg, textureless areas and non-Lambertian surfaces, which are common in human-made scenes~\cite{schops2017multi}. 
Therefore, using prior information to guide the MVS algorithm in these challenging regions is a promising research direction. 
We elaborate 4 typical examples of prior assistance as follows. 

\noindent\textbf{Monocular Depth:} Recently, significant progress has been made in monocular depth estimation~\cite{yang2024depth,yang2024depthv2}, with existing methods achieving accurate results on diverse in-the-wild datasets. Leveraging monocular depth as a strong prior can enhance robustness in challenging scenarios such as low-textured regions and reflective surfaces~\cite{yu2022monosdf,xu2025depthsplat}. For example, monocular depth can be used to regularize or inpaint depth estimation in these challenging regions. %

\noindent \textbf{Surface Normal:} As a non-local representation of the geometry compared with the depth map, surface normal has been proven effective in improving reconstruction performance in recent works~\cite{kusupati2020normal,liu2020depth}. %
Similar to monocular depth~\cite{yang2024depth,yang2024depthv2}, monocular normal estimation~\cite{eftekhar2021omnidata} produces high-quality normal priors on in-the-wild data, which can improve reconstruction quality in low-textured regions\cite{yu2022monosdf,wu2024gomvs}.

\noindent \textbf{Shape Prior:} %
For indoor scenes where common textureless areas, \eg, walls, planes are suitable choices of the geometric primitives and are exploited in traditional methods~\cite{xu2020acmp}. 
Recently, PlaneMVS~\cite{liu2022planemvs} and PlanarRecon~\cite{xie2022planarrecon} explicitly estimate plane parameters for depth refinement or holistic reconstruction. 
In addition, predicting object-level attributes simultaneously when estimating depth~\cite{osman2017semantic} is also a feasible solution for specific applications like urban modeling.

\noindent \textbf{Semantic Segmentation:} Intuitively, points assigned with the same semantic labels may be more likely to lie on the same 3D plane.
Some attempts~\cite{stathopoulou2019semantic,stathopoulou2021semantically} employ a rule-based protocol to utilize semantic segmentation for better matching and depth quality. 
Manhattan-SDF~\cite{guo2022neural} relies on 2D segmentation results to enforce Manhattan world priors to handle low-textured planar regions.
Similarly, Shvets~\etal~\cite{shvets2024joint} also employ SAM (Segment Anything Model) encoder to generate semantic features to enhance MVS.

\section{Conclusion}\label{sec:conclusion}
In this survey, we have provided a comprehensive review of the learning-based MVS methods, which are categorized into: depth map-based, voxel-based, NeRF-based, 3D Gaussian Splatting-based and large feed-forward methods. Particularly, we have devoted significant attention to depth map-based methods because of their conciseness, flexibility and scalability. We explain key aspects such as datasets utilized, general working pipelines, and algorithmic intricacies. Furthermore, we have summarized and compared the quantitative performance of these methods across popular benchmarks, providing valuable insights into their efficacy and applicability. Finally, our discourse extends to an in-depth exploration of potential research directions for the future of MVS, highlighting avenues for further investigation and innovation in the field.

\begin{center}
      {\Large \bf Supplementary Material}
\end{center}
\setcounter{section}{0}
\section{Preliminaries}

\subsection{Plane Sweep}

\begin{figure}[!h]
    \centering
    \includegraphics[width=0.7\linewidth]{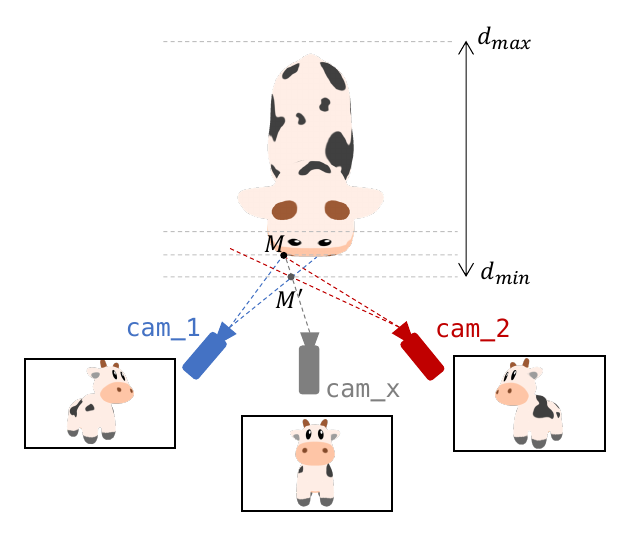}
    \caption{Illustration of plane sweep algorithm~\cite{collins1996space}. To estimate the depth map for reference image (\textit{cam\_x}), neighboring source images (\textit{cam\_1}, \textit{cam\_2}) are projected with homography to fronto-parallel planes of the reference view frustum.}
    \label{fig:sweep}
\end{figure}

We take a toy example of three cameras viewing the same object as illustrated in~\cref{fig:sweep}. 
Plane sweep is then performed on the frustum of \textit{cam\_x} (termed as the reference camera) by creating a series of fronto-parallel hypothesized planes within a given range (typically $[d_{\textrm{min}},d_{\textrm{max}}]$), with each plane corresponding to a depth value \wrt \textit{cam\_x}. 
Let's then examine two 3D points, $M$ and $M'$, as examples of occupied and unoccupied positions on hypothesized planes, respectively. 
For $M$, all of the three cameras capture the identical point lying on the geometry surface with photometric consistency. 
In contrast, the photometric consistency of the observations of $M'$ are poor, indicating that $M'$ is an invalid hypothesis. 
To evaluate the similarity of the observations towards $M$ and $M'$, the images of \textit{cam\_1} and \textit{cam\_2} (termed as the source cameras) are warped to each plane by homography. 
These warped images are then compared against the scaled images of \textit{cam\_x}. Depth hypotheses with high similarity measures are considered reliable.

Regarding selecting $D$ depth hypotheses from the triangulated depth range $[d_{\textrm{min}},d_{\textrm{max}}]$, there are two main schemes, namely the forward depth sampling and the inverse depth sampling.
The naive forward sampling divides the depth range into $D-1$ depth intervals with identical lengths.
Given a depth index $k$, we have
\begin{equation}
    d_k = d_{\textrm{min}} + \frac{k}{D-1}(d_{\textrm{max}}-d_{\textrm{min}}), k=0,\ldots,D-1,
\end{equation}
where the depth hypotheses distribute uniformly between the two ends.
While for the inverse sampling scheme~\cite{xu2020learning_inverse}, we sample uniformly in the multiplicative inverse of $d$, such that
\begin{equation}
    \frac{1}{d_k} = \frac{1}{d_{\textrm{max}}}+ \frac{k}{D-1} (\frac{1}{d_{\textrm{min}}} - \frac{1}{d_{\textrm{max}}}), k=0,\ldots,D-1.
\end{equation}
In this way, the distribution of sampled depth values becomes sparser with $d$ approaching to $d_{\textrm{max}}$. 
When reconstructing unbounded outdoor scenes, where the depth range is rather large, the inverse sampling will be a reasonable choice since it samples more densely at the foreground.
In addition, as revealed in \cite{xu2022learning}, the inverse sampling leads to a uniform sampling on the projected epipolar lines of source images.

\subsection{Depth Fusion}
To remove outliers and improve reconstruction quality, offline MVS methods filter out outliers before converting the depth maps to point clouds. 
Depth filtering mainly involves photometric and geometric filtering~\cite{yao2018mvsnet}. 

For photometric consistency filtering, a per-pixel confidence is estimated to measure the confidence of depth estimation, \ie, probability that the ground truth depth is within a small range near the estimation. Then a threshold can be set to filter depth values with low confidence. 
To estimate this confidence, existing methods either follow MVSNet~\cite{yao2018mvsnet} and estimate photometric confidence from the probability volume~\cite{yao2019recurrent,yan2020dense,wei2021aa, yao2018mvsnet, gu2020cascade, wang2021patchmatchnet} or use a network to predict the confidence of estimation~\cite{wang2021itermvs, wang2025lightweight}. 

For geometric consistency filtering, the consistency of depth estimations are measured among multiple views. %
For a pixel $\mathbf{p}$ in the reference view 0, we project it to pixel $\mathbf{p}'$ in its $i$-th neighboring view through its depth prediction $d_0(\mathbf{p})$. 
After looking up the depth for $\mathbf{p}'$, $d_i(\mathbf{p}')$, we re-project $\mathbf{p}'$ back to the reference view at pixel $\mathbf{p}''$ and look up its depth $D_0(\mathbf{p}'')$. 
We consider pixel $\mathbf{p}$ and its depth as consistent to the $i$-th neighboring view, if the distances, in image space and depth, between the original estimate and its re-projection satisfy: 
\begin{equation}
    \xi_d = \| \mathbf{p}-\mathbf{p}''\|_2\leq \tau_1,
\end{equation}
\begin{equation}
   \xi_p = \frac{\|d_0(\mathbf{p}'')-d_0(\mathbf{p})\|}{d_0(\mathbf{p})}\leq \tau_2,
\end{equation}
where $\tau_1$ and $\tau_2$ are thresholds. 
The pixels are considered to be reliable estimations if they are consistent in at least $N$ neighboring views. 
Recently, instead of using predefined thresholds for $\tau_1$ and $\tau_2$, dynamic consistency checking~\cite{yan2020dense} is proposed to dynamically aggregate geometric matching error among all views and improve the robustness of reconstruction. 
Specifically, the dynamic multi-view geometric consistency $C_{\textrm{geo}}(\mathbf{p})$ is computed as:
\begin{equation}
   C_{\textrm{geo}}(\mathbf{p}) = \sum_i \exp{(-(\xi_p+ \lambda \xi_d))},
\end{equation}
where $\lambda$ is a weight to balance the reprojection error in two metrics. Then the outliers with $C_{\textrm{geo}}(\mathbf{p})$ smaller than a threshold are filtered.

\subsection{Evaluation Metrics}
\subsubsection{2D Metrics}
We list the commonly used metrics as follows: mean absolute depth error (Abs), mean absolute relative depth error (Abs Rel) %
and inlier ratio with threshold 1.25 ($\delta<1.25$):
\begin{equation}
    \begin{aligned}
    \text{Abs} = & \frac{1}{N} \sum_{\mathbf{p}} |d(\mathbf{p}) -\hat{d}(\mathbf{p})|, \\
    \text{Abs Rel} = & \frac{1}{N} \sum_{\mathbf{p}} \frac{|d(\mathbf{p}) -\hat{d}(\mathbf{p})|}{d(\mathbf{p})}, \\
    \text{Inlier Ratio} = & \frac{1}{N} \sum_{\mathbf{p}} \mathbbm{1} \left[ \frac{d(\mathbf{p})}{1.25} < \hat{d}(\mathbf{p}) < 1.25d(\mathbf{p}) \right], \\
    \end{aligned}
\end{equation}
where $d(\mathbf{p})$ denotes the ground truth depth, $\hat{d}(\mathbf{p})$ denotes the estimation, $N$ denotes the number of pixels with valid depth measurements and $\mathbbm{1}[\cdot]$ denotes the indicator function. 

\subsubsection{3D Metrics}
\noindent \textbf{Precision/Accuracy:} 
Precision/Accuracy measures the percentage of predicted points that can be matched to the ground truth point cloud. Considering a point $\mathbf{P}_p$ in the predicted point cloud, it is considered to have a good match in the ground truth point cloud $\{\mathbf{P}_g\}$ if
\begin{equation}
    \|\mathbf{P}_p - \mathop{\arg\min}\limits_{\mathbf{P}\in \{\mathbf{P}_g\}} \|\mathbf{P}-\mathbf{P}_p\|_2\|_2 \leq \tau,
\end{equation}
where $\tau$ is a scene-dependent threshold assigned by datasets, usually set to a large value for large-scale scenes. 
Note that in some datasets, instead of being measured by percentage~\cite{knapitsch2017tanks,schops2017multi}, precision/accuracy is measured by mean or median absolute distance~\cite{aanaes2016large}. 

\noindent \textbf{Recall/Completeness:}
Recall/Completeness measures the percentage of ground truth points that can be matched to the predicted point cloud. 
For a point $\mathbf{P}_g$ in the ground truth point cloud, it is considered a good match in the predicted point cloud $\{\mathbf{P}_p\}$ if
\begin{equation}
    \|\mathbf{P}_g - \mathop{\arg\min}\limits_{\mathbf{P}\in \{\mathbf{P}_p\}} \|\mathbf{P}-\mathbf{P}_g\|_2\|_2 \leq \tau. 
\end{equation}
Similar to precision/accuracy, recall/completeness can be measured by the percentage of points~\cite{knapitsch2017tanks,schops2017multi} or measured by the absolute distance~\cite{aanaes2016large}, similar to Chamfer distance.

\noindent \textbf{F-Score:} 
The two aforementioned metrics measure the accuracy and completeness of predicted point clouds. However, each of these metrics alone cannot present the overall performance since different MVS methods adopt different assumptions. A stronger assumption usually leads to higher accuracy but lower completeness. 
If only precision/accuracy is reported, it would favor MVS algorithms that only include estimated points of high certainty. On the other hand, if only recall/completeness is reported it would favor MVS algorithms that include everything, regardless of accuracy.
Therefore, F-score, an integrated metric, is introduced~\cite{knapitsch2017tanks,schops2017multi}. F-score is the harmonic mean of precision and recall. It is sensitive to extremely small values and tends to get more affected by smaller values, which means that F-score does not encourage imbalanced results. 
However, in most cases, F-score still suffers from unfairness due to the limitations of ground truth, \eg, sparse and incomplete point clouds may penalize for filling in the areas that are not present in the ground truth~\cite{murez2020atlas}. %

\section{Supervised Method with Depth Representation}
\subsection{Cost Volume Regularization}
\begin{figure*}[t!]
    \centering
    \includegraphics[width=0.9\linewidth]{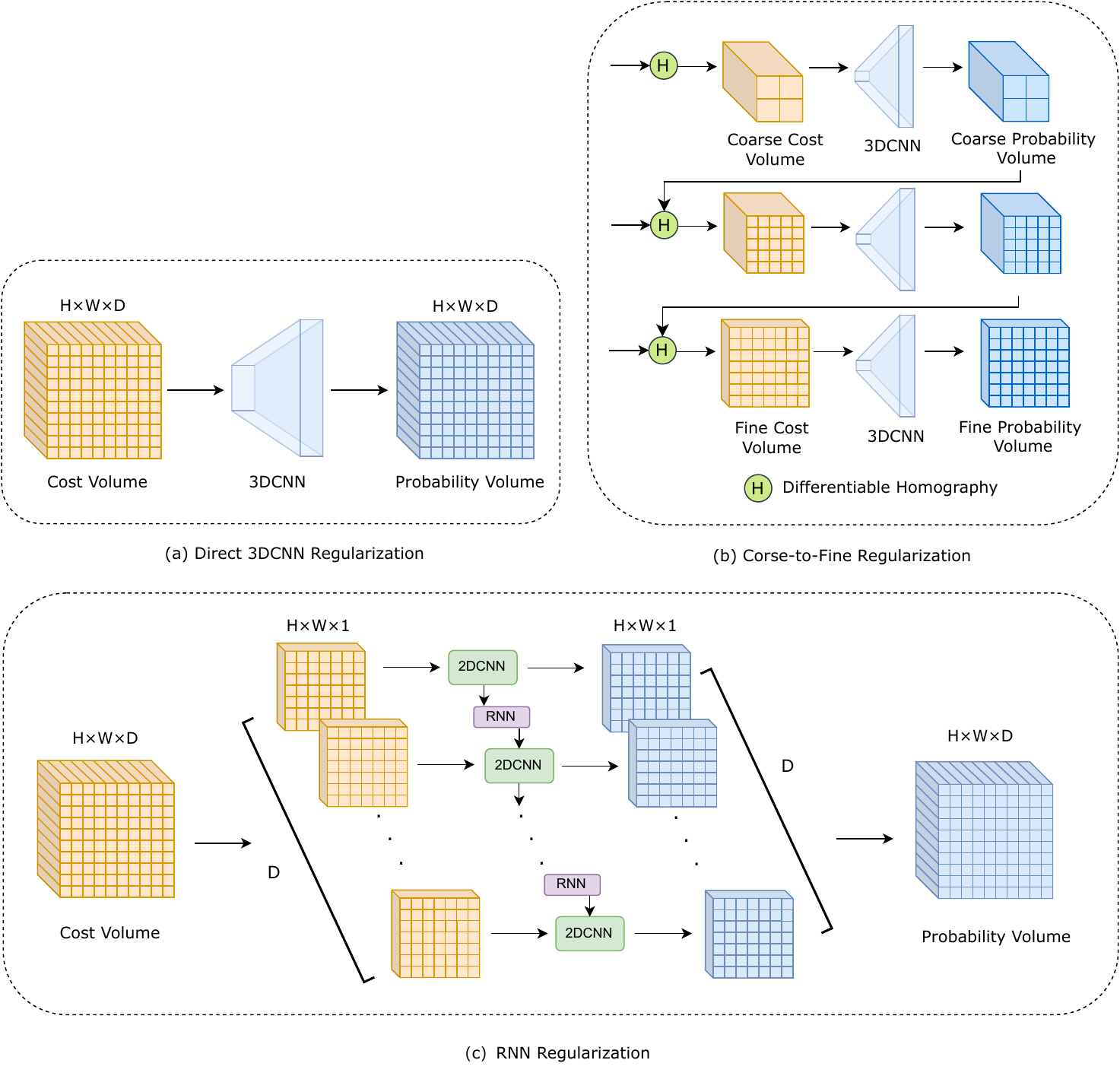}
    \vspace{-3mm}
    \caption{Illustration of typical cost regularization schemes for offline MVS methods that use 4D cost volumes. (a) direct 3D CNN regularization~\cite{yao2018mvsnet,chen2019point,luo2019pmvsnet,xu2020learning_inverse} applies a 3D CNN to aggregate contextual information; (b) coarse-to-fine regularization~\cite{gu2020cascade, cheng2020deep, yang2020cost, xu2022learning} constructs multi-scale cost volumes based on coarse prediction and uses 3D CNN regularization on each scale; (c) RNN regularization~\cite{yao2019recurrent,yan2020dense, wei2021aa} sequentially regularizes 2D slices of the cost volume to reduce memory consumption.}
    \label{fig:regularization}
\end{figure*}

In \cref{fig:regularization}, we visualize the three main categories to perform cost volume regularization for offline MVS methods: direct 3D CNN, coarse-to-fine and RNN. 

\begin{table}[tb!]
    \caption{Quantitative results of voxel-based MVS methods on ScanNet~\cite{dai2017scannet}. %
    }
    \centering
    \begin{tabular}{c|c|c|c}
    \hline
    \textbf{Methods} & Prec. $\uparrow$ & Recall $\uparrow$ & F-score $\uparrow$ \\
    \hline
    Atlas~\cite{murez2020atlas} & 0.565 & 0.598 & 0.578 \\
    NeuralRecon~\cite{sun2021neuralrecon} & 0.684 & 0.479 & 0.562 \\
    TransformerFusion~\cite{bozic2021transformerfusion} & 0.648 & 0.547 & 0.591 \\
    VoRTX~\cite{stier2021vortx} & 0.688 & 0.607 & 0.644 \\
    VisFusion~\cite{gao2023visfusion} & 0.695 & 0.527 & 0.598 \\
    \hline
    \end{tabular}
    \label{tab:voxel-based}
\vspace{-10pt}
\end{table}

\section{Benchmark Performance}

In this section, we summarize the performance of different methods on the common benchmarks they use. Specifically, we summarize the quantitative results of online depth map-based MVS in~\cref{tab:online_mvs}, offline depth map-based MVS in~\cref{tab:offline-mvs}, voxel-based MVS in~\cref{tab:voxel-based}, and NeRF/3D Gaussian Splatting-based MVS in~\cref{tab:nerf_3dgs_dtu}.

\begin{table*}[tb]
    \caption{Quantitative results of online MVS methods on ScanNet~\cite{dai2017scannet} and 7-Scenes~\cite{glocker2013real}. }
    \centering
    \begin{tabular}{c|c|c|c|c|c|c}
    \hline
    \multirow{2}{*}{\textbf{Methods}} & \multicolumn{3}{c}{\textbf{ScanNet~\cite{dai2017scannet}}} & \multicolumn{3}{|c}{\textbf{7-Scenes~\cite{glocker2013real}}} \\
    \cline{2-7}
    & abs $\downarrow$ & abs-rel $\downarrow$ & $\eta < 1.25$ $\uparrow$ & abs $\downarrow$ & abs-rel $\downarrow$ & $\eta < 1.25$ $\uparrow$ \\
    \hline
    MVDepthNet~\cite{wang2018mvdepthnet} & 0.167 & 0.087 & 0.925 & 0.201 & 0.117 & 0.877 \\
    Neural-RGBD~\cite{liu2019neural} & 0.236 & 0.122 & 0.850 & 0.214 & 0.131 & 0.865 \\
    GP-MVS~\cite{hou2019multi} & 0.149 & 0.076 & 0.940 & 0.174 & 0.100 & 0.903 \\
    DeepVideoMVS~\cite{duzceker2021deepvideomvs} & 0.119 & 0.060 & 0.965 & 0.145 &  0.038 & 0.938 \\
    MaGNet~\cite{bae2022multi} & 0.147 & 0.081 & 0.930 & 0.213 & 0.126 & 0.855 \\
    RIAV-MVS~\cite{cai2023riav} & 0.139 & 0.075 & 0.938 & 0.178 & 0.100 & 0.897 \\
    SimpleRecon~\cite{sayed2022simplerecon} & 0.089 & 0.043 & 0.981 & 0.105 & 0.058 & 0.974 \\
    DoubleTake~\cite{sayed2024doubletake} & 0.077 & 0.037 & 0.984 & 0.099 & 0.053 & 0.970 \\
    \hline
    \end{tabular}
    \label{tab:online_mvs}

\end{table*}

\begin{table*}[tb]
    \caption{Quantitative results of point cloud evaluation on MVS benchmarks~\cite{aanaes2016large,knapitsch2017tanks,schops2017multi} for offline MVS methods. ``Finetuned with BlendedMVS" denotes whether methods are finetuned on BlendedMVS dataset~\cite{yao2020blendedmvs} before evaluation on Tanks and Temples~\cite{knapitsch2017tanks} and ETH3D~\cite{schops2017multi}. }
    \centering
    \resizebox{0.98\textwidth}{!}{ 
    \begin{tabular}{c|c|c|c|c|c|c|c|c|c}
    \hline
    \multicolumn{2}{c|}{\multirow{2}{*}{\textbf{Methods}}} & \multicolumn{3}{c|}{\textbf{DTU~\cite{aanaes2016large}}} & Finetuned with &\multicolumn{2}{c|}{\textbf{Tanks and Temples~\cite{knapitsch2017tanks}}} & \multicolumn{2}{c}{\textbf{ETH3D~\cite{schops2017multi}}} \\
    \cline{3-5} \cline{7-10}
    \multicolumn{2}{c|}{} & Acc. $\downarrow$ & Comp. $\downarrow$ & Overall $\downarrow$ & BlendedMVS~\cite{yao2020blendedmvs} & Intermediate $F_1$ $\uparrow$ & Advanced $F_1$ $\uparrow$ & Training $F_1$ $\uparrow$ & Test $F_1$ $\uparrow$ \\
    \hline
    \multirow{6}{*}{Direct 3D CNN} 
    & MVSNet~\cite{yao2018mvsnet} & 0.396 & 0.527 & 0.462 & \xmark & 43.48 & - & - & -\\
    & P-MVSNet~\cite{luo2019pmvsnet} & 0.406 & 0.434 & 0.420 & \xmark & 55.62 & - & - & -\\
    & CIDER~\cite{xu2020learning_inverse} & 0.417 & 0.437 & 0.427 & \xmark & 46.76 & 23.12 & - & - \\
    & PointMVSNet~\cite{chen2019point} & 0.342 & 0.411 & 0.376 & - & - & - & - & - \\
    & PVA-MVSNet~\cite{yi2020pyramid} & 0.379 & 0.336 & 0.357 & \xmark & 54.46 & - & - & - \\
    & Fast-MVSNet~\cite{yu2020fast} & 0.336 & 0.403 & 0.370 & \xmark & 47.39 & - & - & -\\
    \hline
    \multirow{4}{*}{RNN} 
    & R-MVSNet~\cite{yao2019recurrent} & 0.383 & 0.452 & 0.417 & \xmark & 48.40 & 24.91 & - & -\\
    & $D^2$HC-RMVSNet~\cite{yan2020dense} & 0.395 & 0.378 & 0.386 & \xmark & 59.20 & - & - & -\\
    & AA-RMVSNet~\cite{wei2021aa} & 0.376 & 0.339 & 0.357 & \cmark & 61.51 & - & - & - \\
    & BH-RMVSNet~\cite{wei2022bidirectional} & 0.368 & 0.303 & 0.335 & \cmark & 61.96 & 34.81 & - & 79.61 \\
    \hline
    \multirow{22}{*}{Coarse-to-fine} 
    & CasMVSNet~\cite{gu2020cascade} & 0.325 &  0.385 & 0.355 & \xmark & 56.84 & - & - & -\\ 
    & CVP-MVSNet~\cite{yang2020cost} & 0.296 &  0.406 & 0.351 & \xmark & 54.03 & - & - & -\\ 
    & UCS-Net~\cite{cheng2020deep} & 0.338 & 0.349 & 0.344 & \xmark & 54.83 & - & - & -\\
    & AttMVS~\cite{luo2020attention} & 0.383 & 0.329 & 0.356 & \xmark & 60.05 & 37.34 & - & - \\
    & Vis-MVSNet~\cite{zhang2020visibility} & 0.369  & 0.361 & 0.365 & \cmark & 60.03 & - & - & -\\
    & EPP-MVSNet~\cite{ma2021epp} & 0.413 & 0.296 & 0.355 & \cmark & 61.68 & 35.72 &  74.00 & 83.40 \\
    & CDS-MVSNet~\cite{giang2021curvature} & 0.351 & 0.278 & 0.315 & \cmark & 61.58 & - & - & - \\
    & TransMVSNet~\cite{ding2021transmvsnet} & 0.321 & 0.289 & 0.305 & \cmark & 63.52 & 37.00 & - & - \\
    & GBi-Net~\cite{mi2021generalized} & 0.315 & 0.262 & 0.289 & \cmark & 61.42 & 37.32 & - & -\\
    & UniMVSNet~\cite{peng2022rethinking} & 0.352 & 0.278 & 0.315 & \cmark & 64.36 & 38.96 & - & -\\
    & NP-CVP-MVSNet~\cite{yang2022non} & 0.356 & 0.275 & 0.315 & \cmark & 59.64 & - & - & - \\
    & MVSTER~\cite{wang2022mvster} & 0.340 & 0.266 & 0.303 & \cmark & 60.92 & 37.53 & 72.06 & 79.01 \\
    & PVSNet~\cite{xu2022learning} & 0.337 & 0.315 & 0.326 & \cmark & 59.11 & 35.51 & 76.57 & 82.62 \\
    & IS-MVSNet~\cite{wang2022ismvsnet} & 0.351 & 0.359 & 0.355 & \cmark & 62.82 & 34.87 & 73.33 & 83.15 \\
    & HR-MVSNet~\cite{zhu2022hybrid} & 0.332 & 0.310 & 0.321 & \cmark & 63.12 & 34.27 & - & - \\
    & EPNet~\cite{su2023efficient} & 0.299 & 0.323 & 0.313 & \cmark & 63.68 & 40.52 & 79.08 & 83.72 \\
    & GeoMVSNet~\cite{GeoMVSNet} & 0.331 & 0.259 & 0.295 & \cmark & 65.89 & 41.52 & - & - \\ 
    & RA-MVSNet~\cite{zhang2023multi} & 0.326 & 0.268 & 0.297 & \cmark & 65.72 & 39.93 & - & - \\ 
    & DMVSNet~\cite{ye2023constraining} & 0.338 & 0.272 & 0.305 & \cmark & 64.66 & 41.17 & - & - \\
    & ET-MVSNet~\cite{Liu_2023_ICCV} & 0.329 & 0.253 & 0.291 & \cmark & 65.49 & 40.41 & - & - \\
    & GoMVS~\cite{wu2024gomvs} & 0.347 & 0.227 & 0.287 & \cmark & 66.64 & 43.07 & 79.16 & 85.91 \\
    & MVSFormer++~\cite{cao2024mvsformer++} & 0.309 & 0.252 & 0.281 & \cmark & 67.03 & 41.70 & - & 82.99 \\
    \hline
    \multirow{8}{*}{Iterative update} 
    & PatchmatchNet~\cite{wang2021patchmatchnet} & 0.427 & 0.277 & 0.352 & \xmark & 53.15 & 32.31 & 64.21 & 73.12 \\
    & PatchMatch-RL~\cite{lee2021patchmatch} & - & - & - & \cmark & 51.80 & 31.80 & 67.80 & 72.40 \\
    & IterMVS~\cite{wang2021itermvs} &  0.373 & 0.354 & 0.363 & \cmark & 56.94 & 34.17 & 71.69 & 80.09\\
    & Effi-MVS~\cite{wang2022efficient} & 0.321 & 0.313 & 0.317 & \xmark & 56.88 & 34.39 & - & - \\ 
    & CER-MVS~\cite{ma2022multiview} & 0.359 & 0.305 & 0.332 & \cmark & 64.82 & 40.19 & - & - \\
    & IGEV-MVS~\cite{xu2023iterative} & 0.331 & 0.326 & 0.324 & - & - & - & - & - \\
    & DiffMVS~\cite{wang2025lightweight} &  0.318 & 0.297 & 0.308 & \cmark & 63.39 & 39.69 & 74.86 & 82.10 \\
    & CasDiffMVS~\cite{wang2025lightweight} &  0.310 & 0.286 & 0.298 & \cmark & 65.87 & 41.81 & 76.76 & 85.11 \\
    \hline
    \end{tabular}
    }
    \label{tab:offline-mvs}
\end{table*}

\begin{table*}[tb!]
    \caption{Quantitative results of NeRF-based and 3D Gaussian Splatting-based MVS methods on DTU~\cite{aanaes2016large}. }
    \centering
    \begin{tabular}{c|ccccc}
    \hline
    \textbf{Methods} & VolSDF~\cite{yariv2021volume} & NeuS~\cite{wang2021neus} & NeuralWarp~\cite{darmon2021deep} & RegSDF~\cite{zhang2022critical} & PET-NeuS~\cite{wang2023pet} \\
    \hline
    Overall $\downarrow$ & 0.86 & 0.87 & 0.68 & 0.72 & 0.71 \\
    \hline
    \hline
    \textbf{Methods} & Geo-NeuS~\cite{fu2022geo} & NeuS2~\cite{wang2023neus2} & PermutoSDF~\cite{rosu2023permutosdf} & Neuralangelo~\cite{li2023neuralangelo} & UniSDF~\cite{wang2023unisdf} \\
    \hline
    Overall $\downarrow$ & 0.51 & 0.70 & 0.68 & 0.61 & 0.64 \\
    \hline
    \hline
    \textbf{Methods} & SuGaR~\cite{guedon2023sugar} & 2DGS~\cite{huang20242d} &  GaussianSurfels~\cite{dai2024high} & GOF~\cite{yu2024gaussian} & PGSR~\cite{chen2024pgsr} \\
    \hline
    Overall $\downarrow$ & 1.33 & 0.80 & 0.88 & 0.74 & 0.49 \\
    \hline
    \end{tabular}
    \label{tab:nerf_3dgs_dtu}
\vspace{-10pt}
\end{table*}

\ifCLASSOPTIONcaptionsoff
  \newpage
\fi

\clearpage
\clearpage
\bibliographystyle{IEEEtran}
\bibliography{egbib}

\begin{IEEEbiography}[{\includegraphics[width=1in,height=1.25in,clip,keepaspectratio]{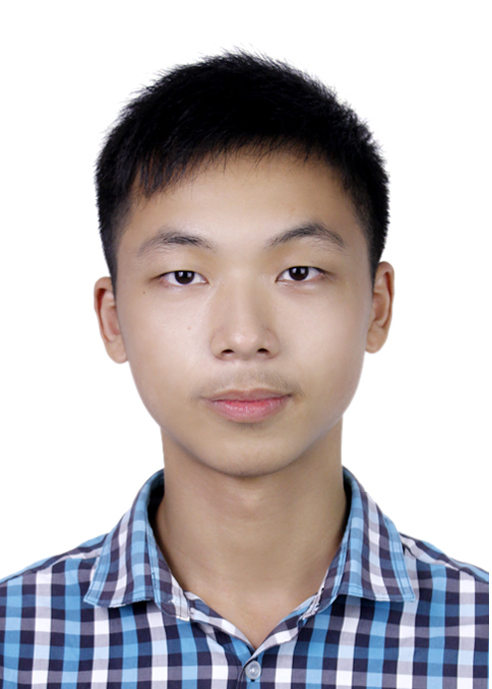}}]
{Fangjinhua Wang} is a Postdoctoral researcher in Computer Science at Computer Vision and Geometry Group, ETH Zurich. He obtained a PhD in Computer Science in the same group. He previously received the MSc degree in Robotics at ETH Zurich. He has a broad interest in 3D computer vision and deep learning, including 3D reconstruction, novel view synthesis, visual localization and mapping, and scene understanding. 
\end{IEEEbiography}
\begin{IEEEbiography}
[{\includegraphics[width=1in,height=1.25in,clip,keepaspectratio]{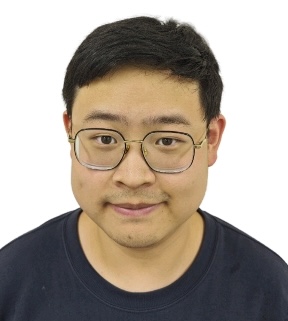}}]{Qingtian Zhu}
is a PhD student at the Graduate School of Information Science and Technology, The University of Tokyo, Japan. Prior to this, he received the MSc degree from the School of Computer Science, Peking University, China. His fields of research are computer vision and computer graphics, including 3D vision, photogrammetry, and implicit representations.
\end{IEEEbiography}
\begin{IEEEbiography}
[{\includegraphics[width=1in,height=1.25in,clip,keepaspectratio]{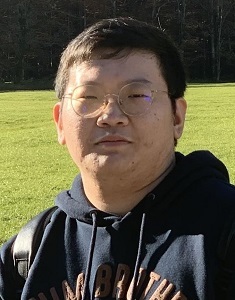}}]
{Di Chang} is currently a PhD student in the Thomas Lord Department of Computer Science and the Institute of Creative Technologies at University of Southern California. His current research interests lie in human-centric computer vision, 3D reconstruction, and generative models.
\end{IEEEbiography}
\vspace{-40pt}
\begin{IEEEbiography}
[{\includegraphics[width=1in,height=1.25in,clip,keepaspectratio]{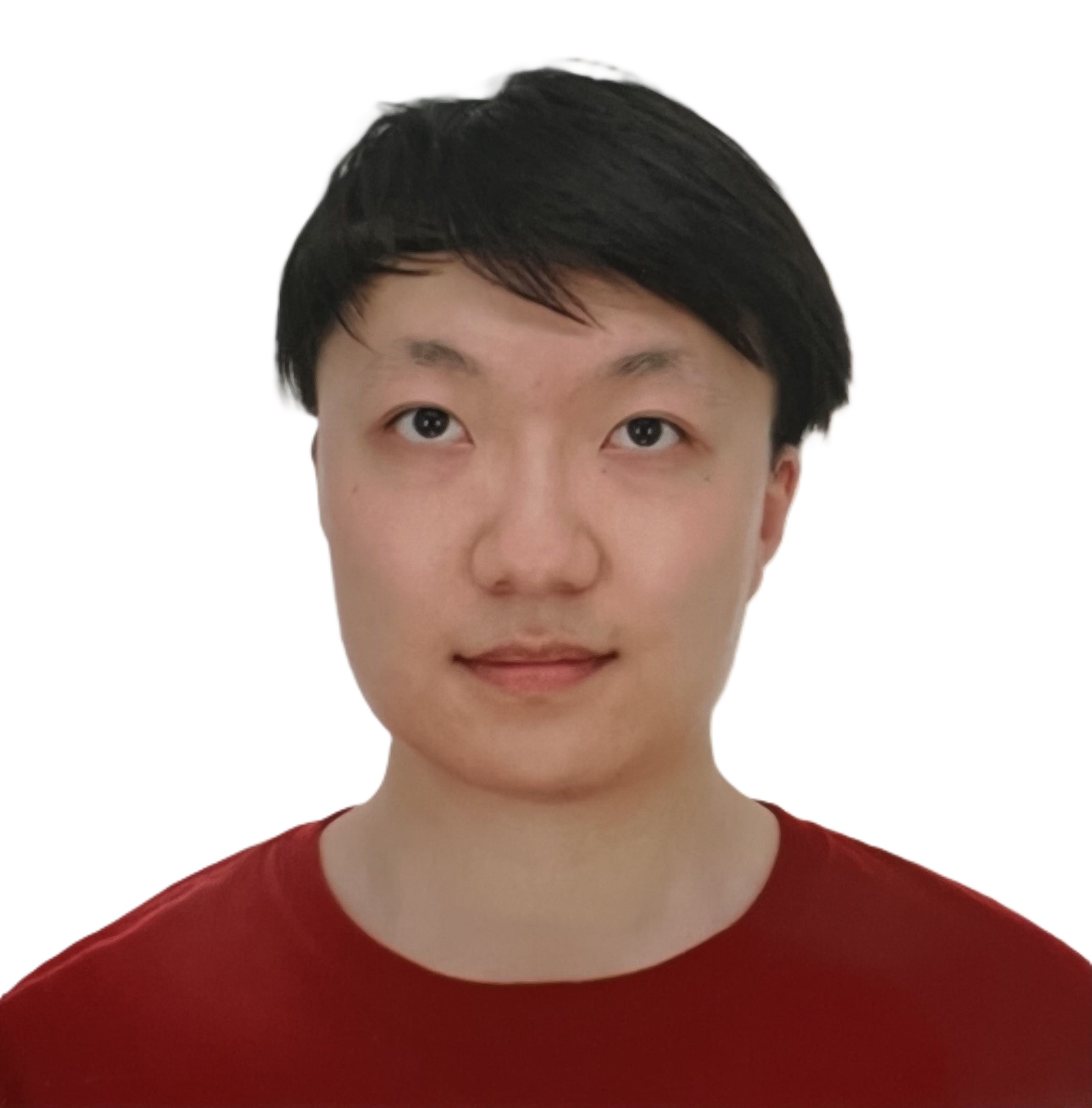}}]
{Quankai Gao} is currently a PhD student in computer science department at University of Southern California. Before that, he earned his MSc degree in computer science at the University of Southern California. His research interests lie at computer vision and computer graphics, including rendering, 3D scene understanding, reconstruction and generation.
\end{IEEEbiography}
\vspace{-40pt}
\begin{IEEEbiography}
[{\includegraphics[width=1in,height=1.25in,clip,keepaspectratio]{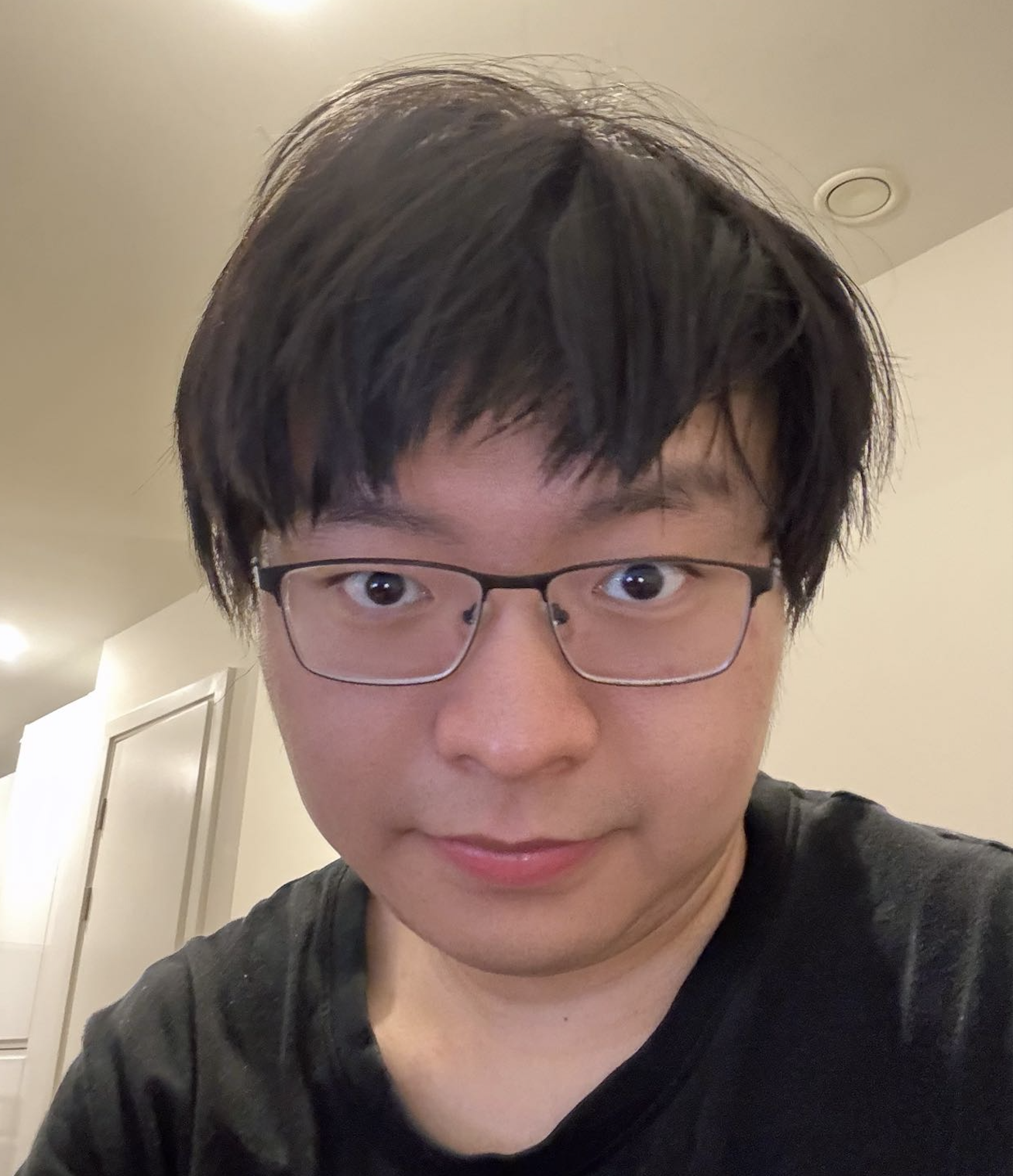}}]
{Junlin Han} is currently a PhD student in the Department of Engineering at the University of Oxford. Before that, he obtained his BSc from the Australian National University. His research interests lie in computer vision and deep learning, including 3D reconstruction, generation, and editing.
\end{IEEEbiography}
\vspace{-40pt}
\begin{IEEEbiography}[{\includegraphics[width=1in,height=1.25in,clip,keepaspectratio]{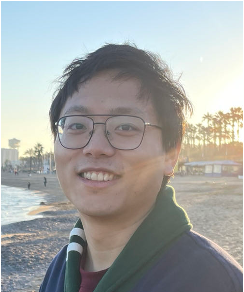}}]
{Tong Zhang} received the Ph.D. degree from the Australian National University in 2020. He is currently a Tenure-Track Assistant Professor at the University of Chinese Academy of Sciences. Previously, he was a Postdoctoral Researcher at IVRL, EPFL. His research interests include subspace clustering, representation learning, and 3D vision. He received the ACCV 2016 Best Student Paper Honorable Mention and the BMVC 2025 Best Poster Award, and was a CVPR 2020 Paper Award Finalist.
\end{IEEEbiography}
\vspace{-40pt}
\begin{IEEEbiography}[{\includegraphics[width=1in,height=1.25in,clip,keepaspectratio]{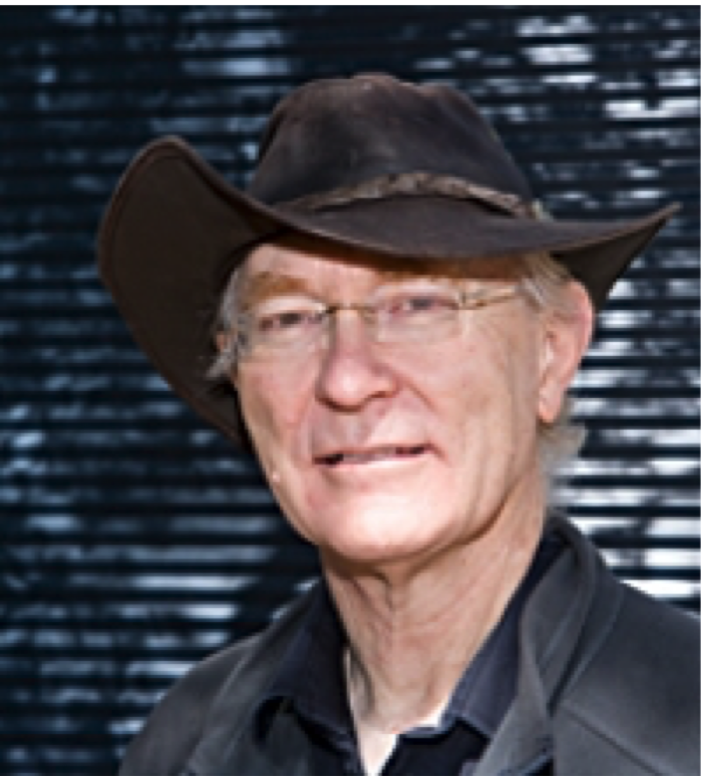}}]
{Richard Hartley} is a Distinguished Professor Emeritus at the Australian National University, where he has worked since 2001. He also led the Autonomous Systems and Sensor Technology Program at National ICT Australia (now Data61, CSIRO). He is an author of the book Multiple View Geometry in Computer Vision. He is a Fellow of the Royal Society, the Australian Academy of Science, the Australian Mathematical Society, and the IEEE.
\end{IEEEbiography}
\vspace{-40pt}
\begin{IEEEbiography}[{\includegraphics[width=1in,height=1.25in,clip,keepaspectratio]{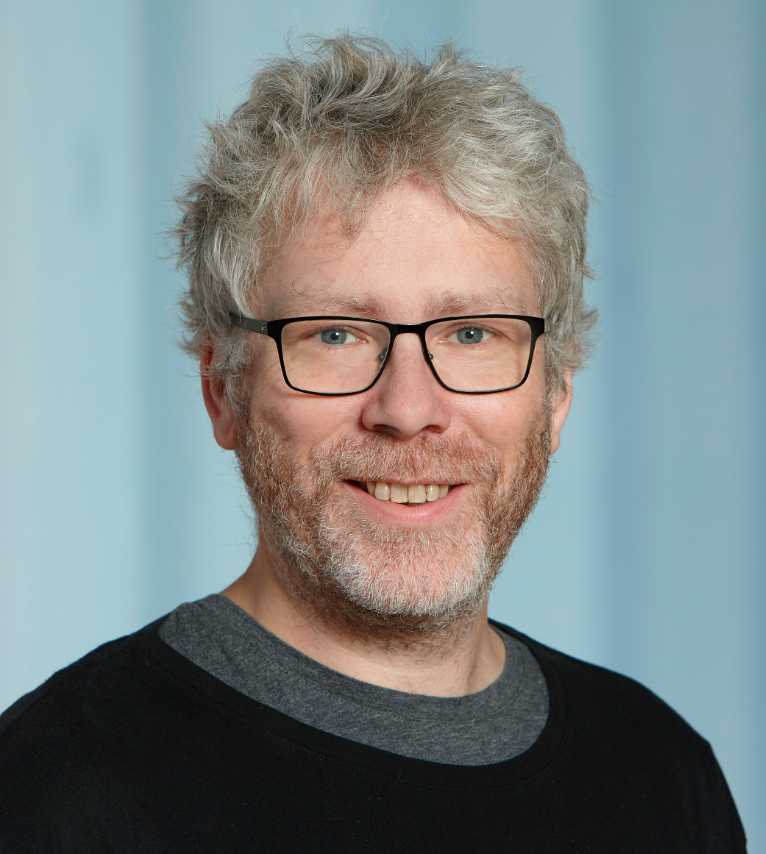}}]{Marc Pollefeys}
is a Prof. of Computer Science at
ETH Zurich and Director of Science at Microsoft.
He is best known for his work in 3D computer
vision, but also for works on robotics, graphics,
machine learning, and camera-based self-driving
cars and drones. He received a M.Sc. and a PhD
from the KU Leuven in Belgium in 1994 and 1999,
respectively. He became an assistant professor
at the University of North Carolina in Chapel Hill
in 2002 and joined ETH Zurich as a full professor
in 2007.
\end{IEEEbiography}

\end{document}